\documentclass[10pt,twocolumn,letterpaper]{article}

\usepackage[pagenumbers]{cvpr}
\usepackage{colortbl}
\usepackage{tabularx}
\usepackage{balance}
\definecolor{GroupGray}{RGB}{240,240,240}
\definecolor{OursBlue}{RGB}{225,239,255}

\definecolor{cvprblue}{rgb}{0.21,0.49,0.74}
\usepackage[breaklinks,colorlinks,allcolors=cvprblue]{hyperref}

\def\confName{CVPR}
\def\confYear{2027}

\title{Look at What Matters: Learning Lesion Prototypes from Severity Labels for Domain-Generalized Diabetic Retinopathy Grading}

\author{
Ziyang Zhang$^{1}$ \quad Juming Xiong$^{1}$ \quad Yanfan Zhu$^{1}$ \quad Yuechen Yang$^{1}$ \\
Junchao Zhu$^{1}$ \quad Zhengyi Lu$^{1}$ \quad Chongyu Qu$^{1}$ \quad Yiming Zhong$^{1}$ \\
Yalin Zheng$^{2}$ \quad Yuankai Huo$^{1}$ \quad He Zhao$^{2}$ \\
$^{1}$Vanderbilt University \qquad $^{2}$University of Liverpool \\
{\tt\small ziyang.zhang@vanderbilt.edu, he.zhao@liverpool.ac.uk}
}

\begin{document}
\maketitle
\begin{abstract}
Deep learning models for diabetic retinopathy (DR) grading often generalize poorly to unseen imaging domains, in part because predictions may exploit correlations specific to source datasets rather than retinal pathology. We address this challenge by constructing grading decisions from lesion evidence learned using only image severity labels. Clinical implications of DR grades provide multiple instance constraints that guide competitive learning of lesion and background prototypes at multiple scales. An additive ordinal scoring model then predicts severity exclusively from individual lesion prototype responses, explicitly connecting learned evidence with grading decisions without a global feature classification pathway. To improve the reliability of this evidence, we introduce teacher guided preservation under image degradations and spatial regularization against persistent coordinate preferences. Across six public DR datasets under a leave one domain out protocol, our method attains a mean quadratic weighted kappa of $0.761$, outperforming the strongest competing method by $0.036$. Localization and prototype removal analyses show that the learned responses align with retinal lesions and contribute to grading decisions. Experiments across five other prototype architectures and under image degradations further demonstrate the generality and robustness of the approach.
\end{abstract}
    
\section{Introduction}
\label{sec:intro}

Diabetic retinopathy (DR) is a leading cause of vision impairment among working-age adults worldwide~\cite{diabetic}. Reliable automated grading from color fundus photographs is therefore important for large-scale screening and clinical assessment. Despite substantial progress in automated DR grading, models trained on available datasets often generalize poorly to unseen imaging domains. Differences in imaging devices, acquisition conditions, preprocessing, and patient populations introduce distribution shifts that challenge model generalization. Domain generalization (DG) aims to address this challenge by learning models that generalize to unseen domains without access to their data during training.

Existing DG approaches seek to improve generalization through feature diversification~\cite{mixstyle} and flat-minima optimization~\cite{swad}. For DR grading, methods such as GDRNet~\cite{gdrnet}, AdvBlur~\cite{advblur}, and GAD~\cite{gad} address domain shifts through augmentation and domain-aware learning. Prototype-based representations offer another approach to capturing transferable patterns. For instance, DECO~\cite{deco} employs class and domain prototypes to improve the robustness of disentangled representations. Other methods incorporate retinal knowledge, with LoASP~\cite{loasp} leveraging structural priors and the uncertainty-aware ordinal grading framework of El Bellaj et al.~\cite{evidord} employing lesion-query attention and evidential ordinal learning. We refer to the latter as EvidOrd throughout this paper. However, learning generalizable representations, including prototype-based features, does not necessarily ensure that grading decisions rely on genuine retinal lesions rather than non-pathological cues correlated with source-domain labels.

\begin{figure}[t]
    \centering
    \includegraphics[width=\columnwidth]{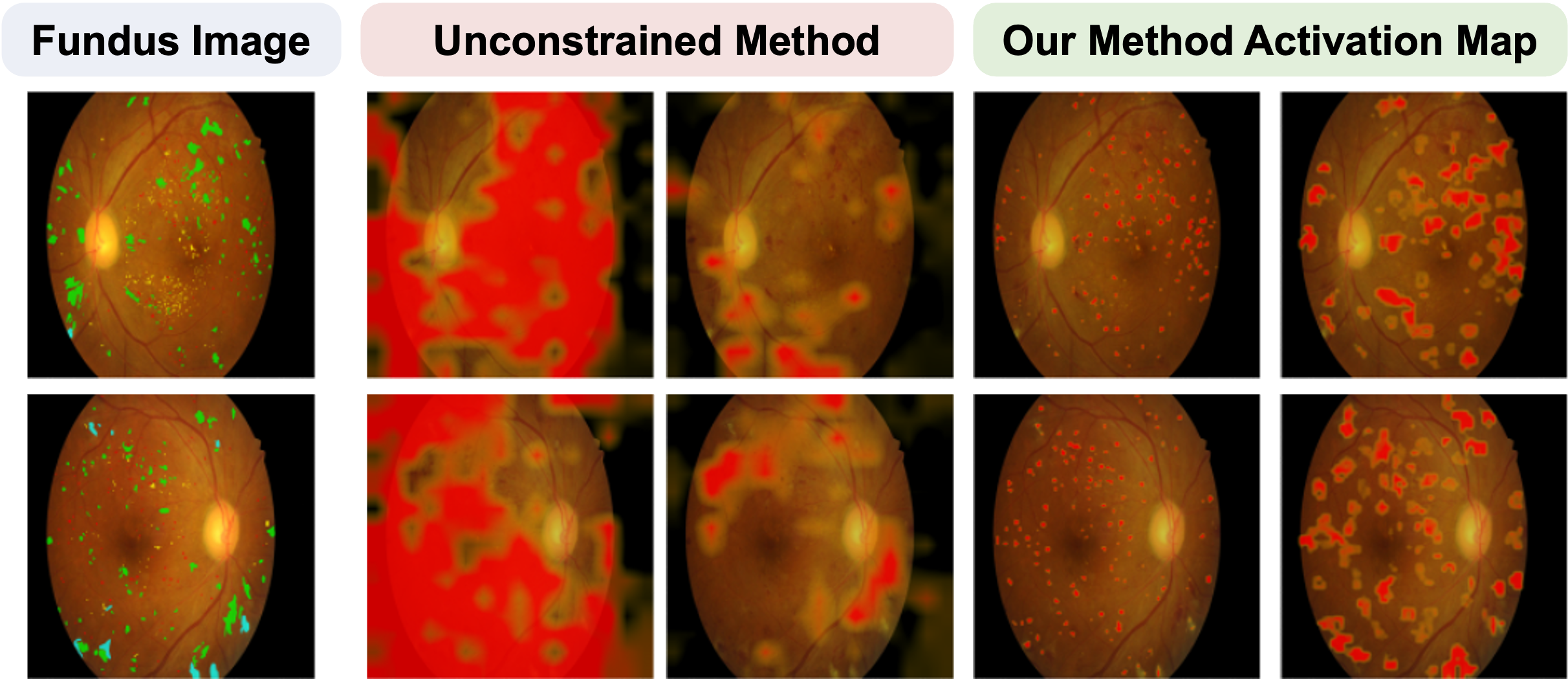}
    \caption{Illustrative motivation for lesion grounded prototype learning on unseen domains. Unconstrained prototype responses can be diffuse, whereas our learned responses are more localized around retinal lesions in these examples. The two most activated prototypes are shown for each method. Pixel annotations are highlighted only for visualization and are never used for training.}
    \label{fig:motivation}
\end{figure}

For DR grading, this distinction is particularly important because disease severity is clinically defined by identifiable retinal abnormalities. The International Clinical Diabetic Retinopathy Severity Scale~\cite{icdr} characterizes severity through the presence, type, and extent of pathological findings, making lesion-associated evidence a meaningful target for domain-generalized representation learning. As illustrated in Fig.~\ref{fig:motivation}, prototypes without pathological guidance may exhibit diffuse responses beyond actual lesion regions.

Obtaining lesion annotations requires substantial expert effort, particularly across diverse imaging domains. DR severity labels offer an alternative source of supervision because they implicitly encode lesion presence. Prior work has exploited this signal through lesion-aware transformers~\cite{lat} and weakly supervised lesion priors~\cite{weaklesion}. However, identifying regions associated with pathology does not by itself ensure that grading decisions rely on them: a classifier can still exploit unrelated visual correlations. We therefore investigate whether constructing severity predictions directly from learned lesion evidence can improve generalization to unseen imaging domains.

\begin{figure*}[t]
    \centering
    \includegraphics[width=\textwidth]{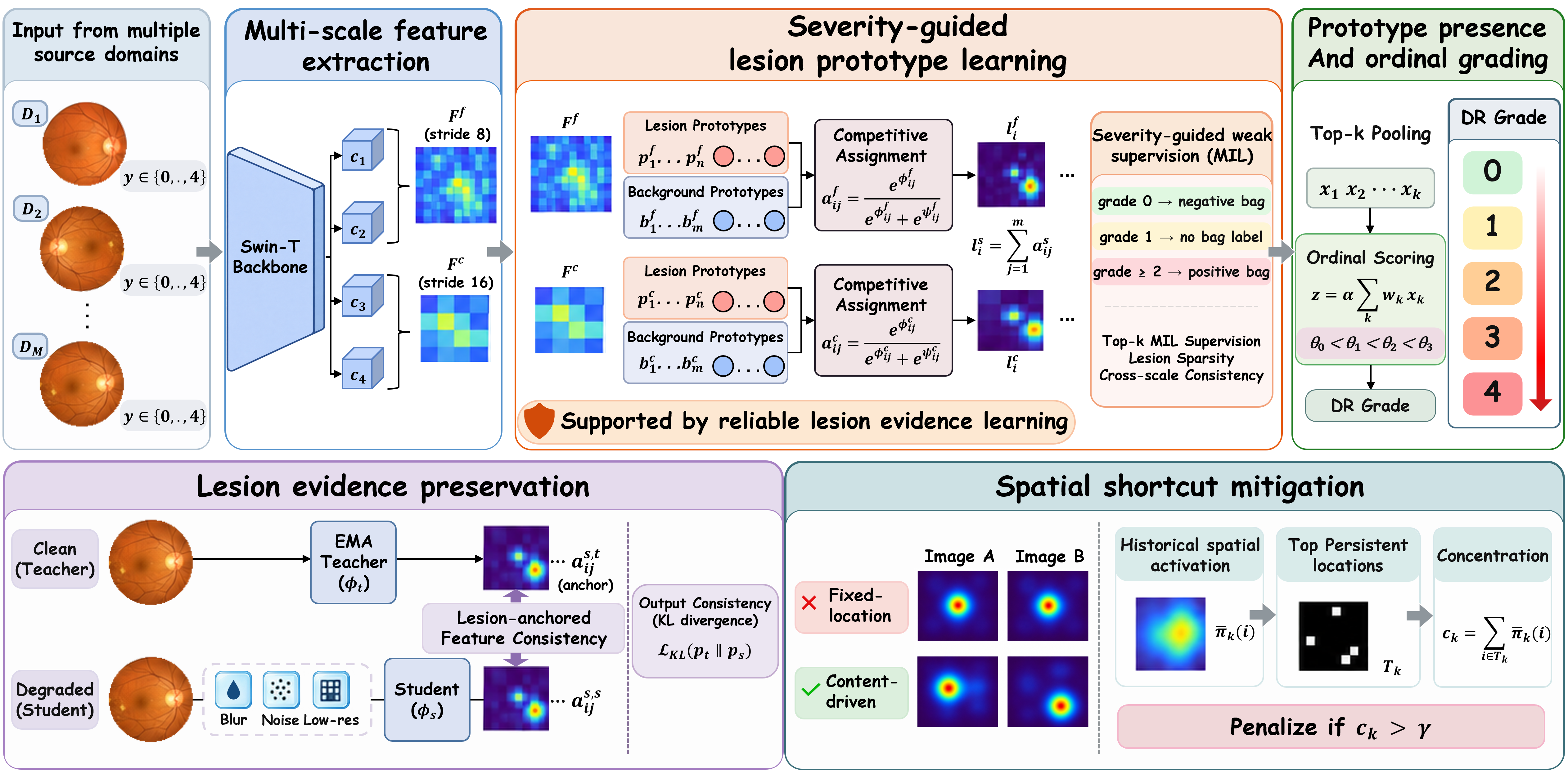}
    \caption{Overview of the proposed LWM framework for domain-generalized DR grading.}
    \label{fig:overview}
\end{figure*}

To address these limitations, we propose Look at What Matters (LWM), a framework that learns lesion and background prototypes at multiple scales from image-level severity labels, without predefined lesion categories or pixel annotations (Fig.~\ref{fig:overview}). Severity-derived multiple instance constraints guide competitive assignments between the two prototype groups. Crucially, instead of allowing an independent classifier to use global features, we construct grading predictions exclusively from individual lesion prototype responses through an additive ordinal scoring model. This architecture makes the learned evidence the only input to the grading decision, while lesion localization is evaluated empirically rather than assumed.

Even with a prototype-based decision path, pathological evidence may remain sensitive to imaging conditions and source-specific shortcuts. Blur, noise, and reduced resolution can obscure subtle lesions, while image-level supervision does not specify where prototypes should respond. Consequently, prototypes may develop preferences for fundus boundaries or fixed image coordinates instead of consistently following pathology. We address these limitations with lesion evidence preservation, which stabilizes prototype representations under image degradations, and spatial shortcut mitigation, which discourages persistent coordinate preferences.

We evaluate the framework on six public DR datasets under a leave-one-domain-out protocol. In addition to grading performance, we assess whether prototype responses localize retinal lesions, whether they influence decisions, whether the supervision transfers to other prototype architectures, and whether learned evidence remains stable under image degradations.

Our main contributions are summarized as follows:
\begin{itemize}
    \item We introduce a lesion grounded framework for DR domain generalization. Severity-derived multiple instance constraints guide competitive lesion and background prototypes, and an additive ordinal scoring model constructs grading predictions directly from their responses without lesion annotations.

    \item We improve the reliability of learned evidence through lesion evidence preservation under image degradations and spatial shortcut mitigation against persistent coordinate preferences.

    \item Across six public DR datasets, we achieve leading generalization performance and show that learned prototypes localize retinal lesions and contribute to grading decisions. Experiments across five existing prototype architectures and under image degradations further evaluate generality and robustness.

\end{itemize}

\section{Method}
\label{sec:method}

\paragraph{Problem formulation.}
We consider domain generalization for diabetic retinopathy (DR) grading into five severity levels. Let $\mathcal{D}_S=\{\mathcal{D}_1,\ldots,\mathcal{D}_M\}$ denote $M$ source domains, where $\mathcal{D}_m=\{(I_i^m,y_i^m)\}_{i=1}^{N_m}$ contains fundus images $I_i^m$ and their severity labels $y_i^m\in\{0,\ldots,4\}$. Training uses only severity labels for each image, without lesion locations, segmentation masks, or lesion type annotations. Our objective is to learn a grading model $f_\theta$ that generalizes to an unseen target domain $\mathcal{D}_T$ without access to target domain data during training.

\paragraph{Framework overview.}
As illustrated in Fig.~\ref{fig:overview}, LWM first learns candidate pathological evidence and then constructs grading predictions entirely from that evidence. Severity-derived multiple instance constraints guide competitive lesion and background prototype learning at two spatial scales. Individual lesion prototype responses form an evidence vector that is the sole input to an additive ordinal grading module; no independent global feature classifier can bypass this path. Lesion evidence preservation and spatial shortcut mitigation further encourage stable, spatially reliable responses. Training requires only image severity labels, and inference uses a single prediction path.

\subsection{Severity Guided Lesion Prototype Learning}
\label{sec:proto}

\paragraph{Feature extraction at multiple scales.}
DR lesions vary substantially in size, motivating representations that capture both fine pathological details and broader retinal context. Given a fundus image $I\in\mathbb{R}^{3\times1024\times1024}$, a Swin-T backbone~\cite{swin} extracts four hierarchical features $c_1,\ldots,c_4$ at strides of 4, 8, 16, and 32. We fuse adjacent stages to construct representations at fine and coarse resolutions:
\begin{equation}
\begin{aligned}
F^f &= \phi_f\big([\mathrm{pool}_2(c_1),c_2]\big),\\
F^c &= \phi_c\big([c_3,\mathrm{up}_2(c_4)]\big),
\end{aligned}
\label{eq:features}
\end{equation}
where $\phi_f$ and $\phi_c$ are $1\times1$ projections to $d=128$ dimensions, producing feature maps at strides 8 and 16, respectively.

\paragraph{Competitive prototype assignment.}
We represent recurring local visual patterns using learnable prototypes without predefined lesion categories. The lesion and background designations describe the roles encouraged by supervision, not annotated pathological classes. At each scale $s\in\{f,c\}$, we maintain lesion and background prototype sets $\mathcal{P}_{\mathrm{les}}^s$ and $\mathcal{P}_{\mathrm{bg}}^s$, indexed by $\mathcal{I}_{\mathrm{les}}^s$ and $\mathcal{I}_{\mathrm{bg}}^s$, respectively, with $p_j^s\in\mathbb{R}^d$. Let $\mathcal{I}^s=\mathcal{I}_{\mathrm{les}}^s\cup\mathcal{I}_{\mathrm{bg}}^s$. Each spatial feature $F_i^s$ is assigned competitively across both prototype groups:
\begin{equation}
a_{ij}^s=\frac{\exp(-\|F_i^s-p_j^s\|_2^2/\tau_s)}{\sum_{j'\in\mathcal{I}^s}\exp(-\|F_i^s-p_{j'}^s\|_2^2/\tau_s)},
\label{eq:assign}
\end{equation}
where $\tau_s$ is a temperature parameter and the denominator spans all prototypes at scale $s$. Since $\sum_j a_{ij}^s=1$, the two prototype groups compete for each spatial feature. Combined with severity supervision, this competition encourages background prototypes to represent normal retinal appearance and lesion prototypes to capture patterns associated with disease. The total lesion assignment at location $i$ is
\begin{equation}
\ell_i^s=\sum_{k\in\mathcal{I}_{\mathrm{les}}^s}a_{ik}^s.
\label{eq:lesion_mass}
\end{equation}

\paragraph{Severity guided multiple instance learning.}
Although lesion annotations are unavailable, DR grades provide clinically meaningful supervision about lesion presence. Images with grade 0 serve as references for normal retinal appearance, whereas images with grades of 2 or above indicate the presence of pathological lesions. We use this distinction to initialize background prototypes from features of healthy images and lesion prototypes from diseased image features that are distant from their nearest healthy image features. We further formulate these constraints within the multiple instance learning (MIL) framework~\cite{mil}, treating each image as a bag of spatial features.

Let $\Omega^s$ denote valid retinal locations at scale $s$. We define the image level lesion response $t^s$ as the mean of the largest $\kappa_s$ values of $\ell_i^s$ over $\Omega^s$, and $\bar{\ell}^s$ as its spatial mean weighted by the fundus mask. Our MIL objective is
\begin{equation}
\begin{aligned}
\mathcal{L}_{\mathrm{bag}}
=\sum_{s\in\{f,c\}}\Big[
&\mathbb{E}_{y=0}(t^s-\delta^-)_{+}\\
+&\mathbb{E}_{y\geq2}(\delta^+-t^s)_{+}\\
+&\mathbb{E}(\bar{\ell}^s-\rho_s)_{+}
\Big],
\end{aligned}
\label{eq:bag}
\end{equation}
where $(u)_+=\max(u,0)$, $\delta^-$ and $\delta^+$ define activation margins, and $\rho_s$ limits average lesion assignment. The first two terms suppress lesion responses in healthy images and enforce positive lesion evidence in diseased images, while the third prevents widespread lesion assignments. Images with grade 1 are excluded from positive and negative bag supervision because their subtle microaneurysms may be unresolved at the feature map resolution, but remain subject to the average mass constraint and grading supervision.

Together, competitive assignment and severity guided MIL encourage lesion responses at a limited set of disease associated locations, while discouraging diffuse activation across the retinal field. To coordinate evidence across scales, we additionally constrain lesion responses at the coarse resolution using their corresponding fine resolution activations:
\begin{equation}
\mathcal{L}_{\mathrm{xs}}=
\frac{1}{|\Omega^c|}
\sum_{i\in\Omega^c}
\left(
\ell_i^c-\max_{j\in\mathcal{C}(i)}\ell_j^f
\right)_+,
\label{eq:cross_scale}
\end{equation}
where $\mathcal{C}(i)$ denotes the $2\times2$ fine resolution locations corresponding to coarse location $i$.

\subsection{Evidence Based Ordinal Grading}
\label{sec:use}

Learning representations associated with retinal lesions alone does not ensure that grading decisions depend on them. We therefore construct severity predictions directly from lesion prototype responses, without an independent classifier operating on global features. Rather than collapsing lesion assignments into a single response, we retain individual prototype activations to represent distinct pathological patterns. For each lesion prototype $k$ at scale $s$, we obtain its presence score $x_k\in[0,1]$ by averaging its largest $\kappa_s$ assignments $a_{ik}^s$ over $\Omega^s$, yielding an evidence vector $\mathbf{x}\in[0,1]^{14}$.

Ordinal regression explicitly models the ordered structure of class labels~\cite{ordinal}. To capture this structure while retaining interpretable contributions from individual lesion prototypes, we construct an additive ordinal scoring model:
\begin{equation}
z=\alpha\sum_{k=1}^{14}w_kx_k,\qquad
P(y>r\mid I)=\sigma(z-\theta_r),
\label{eq:sheet}
\end{equation}

where $r\in\{0,1,2,3\}$, $w_k=\operatorname{softplus}(\hat{w}_k)$, $\alpha$ is a learnable scale that remains positive in our trained models, and $\theta_0<\theta_1<\theta_2<\theta_3$ are ordered thresholds parameterized by positive increments. Class probabilities are recovered from the cumulative predictions. Each prototype contributes an explicit term $\alpha w_kx_k$ to the grading score. Because the grading module receives only the fourteen lesion prototype presence scores, changes in predicted severity with respect to the input image must be mediated by these prototype responses, rather than by a separate global classification pathway. Their individual contributions can therefore be inspected directly; whether the responses align with actual lesions is assessed experimentally.

The grading loss $\mathcal{L}_{\mathrm{grade}}$ is the class weighted cross entropy over the predicted grade probabilities. We additionally regularize the prototype weights with an $\ell_1$ penalty.

\subsection{Reliable Lesion Evidence Learning}
\label{sec:reliable}

Severity supervision provides a basis for learning lesion evidence, but neither guarantees its stability under changes in image quality nor specifies where it should appear. We therefore introduce complementary mechanisms targeting evidence preservation and persistent spatial biases.

\paragraph{Lesion evidence preservation.}
To preserve subtle pathological representations under variations in image quality, we employ a teacher updated through an exponential moving average (EMA)~\cite{meanteacher}. For a subset $\mathcal{B}'$ comprising half of each training batch, the teacher processes the clean image $I$, while the student receives a perturbed view $\tilde I=T_d(I)$. We consider blur, noise, and resolution reduction, which disrupt fine lesion details while preserving spatial correspondence.

Instead of enforcing uniform consistency across the feature map, we use teacher prototype assignments to anchor feature alignment to regions identified by lesion prototypes. For each image $b$ and lesion prototype $k$, we aggregate features from the clean and perturbed views using identical teacher assignments:
\begin{equation}
\begin{aligned}
\mathbf{h}_{bk}^{T}
&=\frac{\sum_{i\in\Omega_b}a_{bik}^{T}F_{bi}^{T}}
{\sum_{i\in\Omega_b}a_{bik}^{T}+\epsilon},\\
\mathbf{h}_{bk}^{S}
&=\frac{\sum_{i\in\Omega_b}a_{bik}^{T}\tilde F_{bi}^{S}}
{\sum_{i\in\Omega_b}a_{bik}^{T}+\epsilon},
\end{aligned}
\label{eq:anchored_features}
\end{equation}
where the feature scale is implicit in $k$, and teacher assignments are treated as fixed aggregation weights. We focus the alignment on sufficiently activated prototypes using $\omega_{bk}=x_{bk}^{T}\mathbb{1}[x_{bk}^{T}>\eta]$, where $\eta$ is a presence threshold. The feature consistency objective is
\begin{equation}
\mathcal{L}_{\mathrm{feat}}
=
\frac{
\sum_{b\in\mathcal{B}'}\sum_k
\omega_{bk}
\left(1-\cos(\mathbf{h}_{bk}^{S},\mathbf{h}_{bk}^{T})\right)
}{
\max\left(1,\sum_{b\in\mathcal{B}'}\sum_k\omega_{bk}\right)
}.
\label{eq:feature_consistency}
\end{equation}

We further align the predicted grade distributions:
\begin{equation}
\mathcal{L}_{\mathrm{out}}
=
\frac{1}{|\mathcal{B}'|}
\sum_{b\in\mathcal{B}'}
\mathrm{KL}\left(
p_{\psi'}(\cdot\mid I_b)
\,\Vert\,
p_{\psi}(\cdot\mid\tilde I_b)
\right),
\label{eq:output_consistency}
\end{equation}
where $\psi'$ and $\psi$ denote the teacher and student parameters, respectively. Alignment of feature representations preserves evidence captured by individual prototypes, while consistency in the output distributions stabilizes the resulting grading predictions. Both objectives operate during training without introducing additional computation during inference.

\paragraph{Spatial shortcut mitigation.}
Supervision derived from severity labels constrains lesion presence without specifying lesion locations, allowing prototypes to exploit nonpathological cues correlated with labels in the source domains. Such cues may include responses outside the retinal field of view (FOV) or persistent preferences for fixed spatial coordinates, whose predictive correlations may not transfer across domains.

We restrict prototype pooling to valid retinal locations $\Omega^s$, estimated using an intensity based fundus mask. During inference, we further retain the largest connected component to exclude disconnected nonretinal regions, while training uses the original intensity based mask.

To suppress bias toward fixed spatial locations, we aggregate each lesion prototype's assignments across the training batch and normalize them over spatial locations to obtain $\pi_k(i)$. Its exponential moving average $\bar{\pi}_k(i)$ tracks persistent coordinate preferences across training iterations. Let $\mathcal{T}_k$ contain the $n_s$ locations with the highest historical activation and define their concentration as $c_k=\sum_{i\in\mathcal{T}_k}\bar{\pi}_k(i)$. We introduce the regularization objective
\begin{equation}
\mathcal{L}_{\mathrm{pos}}
=
\sum_k
\mathbb{1}[c_k>\gamma]
\sum_{i\in\mathcal{T}_k}\pi_k(i),
\label{eq:positional}
\end{equation}
where $\gamma$ is a concentration threshold. By penalizing responses at repeatedly favored coordinates, this objective discourages one form of spatial shortcut across source images without requiring uniform responses within individual images.

\subsection{Training Objective and Model Configuration}
\label{sec:impl}

The overall training objective is
\begin{equation}
\begin{aligned}
\mathcal{L} ={}&
\mathcal{L}_{\mathrm{grade}}
+\mathcal{L}_{\mathrm{bag}}
+0.5\mathcal{L}_{\mathrm{xs}}
+0.05\mathcal{L}_{\mathrm{div}}\\
&+\frac{0.01}{14}\|w\|_1
+\mathcal{L}_{\mathrm{feat}}
+\mathcal{L}_{\mathrm{out}}
+\mathcal{L}_{\mathrm{pos}},
\end{aligned}
\label{eq:objective}
\end{equation}
where $\mathcal{L}_{\mathrm{div}}$ penalizes pairwise cosine similarity among lesion prototypes after centering each prototype bank by the mean of all its prototypes.

We use eight background prototypes per scale, with eight lesion prototypes at the fine resolution and six at the coarse resolution. For MIL supervision, we set $(\kappa_f,\kappa_c)=(16,8)$, $(\delta^-,\delta^+)=(0.2,0.5)$, and $(\rho_f,\rho_c)=(0.10,0.25)$. Optimization details are reported in Sec.~\ref{sec:exp_setup}, with further implementation details in the appendix.
\section{Experiments}
\label{sec:experiments}

\subsection{Experimental Setup}
\label{sec:exp_setup}

\paragraph{Datasets and evaluation protocol.}
We evaluate our framework on six public retinal fundus datasets: FGADR~\cite{fgadr}, APTOS~\cite{aptos}, DDR~\cite{ddr}, IDRiD~\cite{idrid}, Messidor-2~\cite{messidor1, messidor2}, and DeepDRiD~\cite{deepdrid}. 
We standardize their labels into five DR severity grades and adopt a leave one domain out (LODO) evaluation protocol, following established domain generalization evaluation practices~\cite{dg}. In each fold, one dataset serves as the unseen target domain, while the remaining five provide source data for training. Neither target images nor target annotations are used for model training. Our framework requires only image severity labels, without lesion locations, masks, or lesion types. The main grading comparisons, ablations, and degradation experiments are evaluated over three random seeds; the prototype plug-in experiments use one seed per configuration, as specified below.

\paragraph{Baselines and evaluation metrics.}
We compare against ERM and general DG methods, including MixStyle~\cite{mixstyle}, SWAD~\cite{swad}, VREx~\cite{vrex}, Fishr~\cite{fishr}, SAGM~\cite{sagm}, and ERM++~\cite{ermpp}. We also evaluate methods specifically developed for DR classification and grading: DRGen~\cite{drgen}, GDRNet~\cite{gdrnet}, DECO~\cite{deco}, AdvBlur~\cite{advblur}, GAD~\cite{gad}, LoASP~\cite{loasp}, and EvidOrd~\cite{evidord}. All main comparison methods use a Swin-T~\cite{swin} backbone except LoASP, which uses ResNet-50~\cite{resnet}; we additionally report ERM with ResNet-50 for comparison. Quadratic weighted kappa (QWK, $\kappa$) is our primary metric because DR severity is ordinal. We also report accuracy and macro F1.

\paragraph{Implementation details.}
We resize fundus images to $1024\times1024$ and train our Swin-T framework for 15 epochs using AdamW, a batch size of 8, and mixed precision. The learning rate is $2\times10^{-5}$ for the Swin-T backbone and $2\times10^{-4}$ for all remaining trainable modules, including feature projections, prototypes, and the grading head, with a one cycle schedule. The MLP classification head ablation uses the same learning rate of $2\times10^{-4}$ as the additive ordinal head. The EMA teacher has momentum $0.999$ and operates only during training. We select checkpoints by QWK on a validation split drawn from source domains. All methods follow the same LODO protocol and preprocessing pipeline.

\subsection{Comparison with Existing Methods}
\label{sec:exp_main}

\begin{table*}[t]
\centering
\caption{Grading performance across six unseen domains under the LODO protocol. Results are averaged over three random seeds. Best and second best results are highlighted in \textbf{bold} and \underline{underlined}, respectively, including ties at the reported precision. $^{\ast}$Reimplemented in our unified pipeline based on the original paper (public code available). $^{\dagger}$Adapted from official code. $^{\ddagger}$Reimplemented from the original paper without public code. All methods use Swin-T unless otherwise specified.}
\label{tab:lodo}
\small
\setlength{\tabcolsep}{3.5pt}
\renewcommand{\arraystretch}{1.12}
\begin{tabular}{l*{9}{c}}
\toprule
& \multicolumn{6}{c}{Held-out Domain} & \multicolumn{3}{c}{Overall} \\
\cmidrule(lr){2-7}
\cmidrule(lr){8-10}
Method & FGADR & APTOS & DDR & IDRiD & Mess-2 & DeepDRiD & Mean $\kappa$ & Acc & F1 \\
\midrule
\rowcolor{GroupGray}
\multicolumn{10}{l}{\emph{Standard Baselines}} \\
ERM (Swin-T~\cite{swin}) & .462 & .795 & .737 & .745 & .783 & .740 & .710 & .598 & .499 \\
ERM (ResNet-50~\cite{resnet}) & .463 & .831 & .726 & .765 & .776 & .693 & .709 & .604 & .473 \\
\addlinespace[2pt]
\rowcolor{GroupGray}
\multicolumn{10}{l}{\emph{General DG Methods}} \\
MixStyle~\cite{mixstyle}\textsuperscript{$\ast$} & .485 & .794 & .753 & .734 & .818 & .735 & .720 & .603 & .504 \\
SWAD~\cite{swad}\textsuperscript{$\ast$} & .478 & .802 & .731 & .765 & .775 & .737 & .715 & .598 & .500 \\
VREx~\cite{vrex}\textsuperscript{$\ast$} & .454 & .823 & \textbf{.779} & \underline{.785} & .792 & .717 & \underline{.725} & .605 & .488 \\
Fishr~\cite{fishr}\textsuperscript{$\ast$} & .447 & .795 & .738 & .729 & .794 & .724 & .705 & .597 & .501 \\
SAGM~\cite{sagm}\textsuperscript{$\ast$} & .515 & .813 & .745 & .720 & .798 & .737 & .721 & .614 & \underline{.509} \\
ERM++~\cite{ermpp}\textsuperscript{$\ast$} & .464 & .768 & .676 & \textbf{.791} & .662 & .713 & .679 & .559 & .454 \\
\addlinespace[2pt]
\rowcolor{GroupGray}
\multicolumn{10}{l}{\emph{DR-specific Methods}} \\
GDRNet~\cite{gdrnet}\textsuperscript{$\dagger$} & \underline{.525} & .759 & .737 & .780 & .772 & .717 & .715 & .590 & .474 \\
DRGen~\cite{drgen}\textsuperscript{$\ast$} & .471 & .800 & .733 & .749 & .784 & .715 & .709 & .594 & .497 \\
DECO~\cite{deco}\textsuperscript{$\ddagger$} & .458 & .843 & .754 & .755 & .801 & .709 & .720 & .613 & .500 \\
AdvBlur~\cite{advblur}\textsuperscript{$\ddagger$} & .434 & .811 & .746 & .759 & .789 & .708 & .708 & .598 & .495 \\
GAD~\cite{gad}\textsuperscript{$\ddagger$} & .466 & .782 & .726 & .759 & .777 & .735 & .708 & .596 & .499 \\
LoASP~\cite{loasp} (ResNet-50)\textsuperscript{$\ddagger$} & .447 & \textbf{.852} & .757 & .747 & .753 & \underline{.745} & .717 & .612 & .487 \\
EvidOrd~\cite{evidord}\textsuperscript{$\ddagger$} & .449 & .841 & \underline{.777} & .724 & \underline{.819} & .730 & .723 & \underline{.616} & .498 \\
\midrule
\rowcolor{OursBlue}
\textbf{Ours} & \textbf{.551} & \underline{.846} & \underline{.777} & \underline{.785} & \textbf{.849} & \textbf{.755} & \textbf{.761} & \textbf{.618} & \textbf{.511} \\
\bottomrule
\end{tabular}
\end{table*}

\paragraph{Grading performance across unseen domains.}
Table~\ref{tab:lodo} reports grading performance across six unseen domains. Our method achieves the highest mean QWK of $0.761$, exceeding the strongest competing method, VREx ($0.725$), by $0.036$ and the Swin-T ERM baseline ($0.710$) by $0.051$. It also achieves the highest overall accuracy ($0.618$) and macro F1 ($0.511$), indicating that the improvement extends beyond the primary ordinal metric.

The method ranks first on FGADR ($0.551$), Messidor-2 ($0.849$), and DeepDRiD ($0.755$), and second on APTOS ($0.846$), DDR ($0.777$), and IDRiD ($0.785$), including ties at the reported precision. The gains over the strongest baseline reach $0.026$ on FGADR and $0.030$ on Messidor-2. Among methods designed for DR grading, EvidOrd provides the strongest mean QWK ($0.723$), which our method exceeds by $0.038$. These results establish strong performance across heterogeneous target domains. We next examine the contributions of individual components, the pathological relevance and decision role of learned prototypes, and robustness to image degradations.

\subsection{Ablation Studies}
\label{sec:ablation}

\begin{table}[t]
\centering
\caption{Ablation study under the six domain LODO protocol. Results are averaged over three random seeds, with standard deviations reported for QWK. $\Delta\kappa$ is measured relative to the full model. The global feature classifier is the Swin-T ERM baseline from Table~\ref{tab:lodo}.}
\label{tab:ablation}
\small
\setlength{\tabcolsep}{3pt}
\renewcommand{\arraystretch}{1.12}
\begin{tabularx}{\columnwidth}{@{}>{\raggedright\arraybackslash}X c c c r@{}}
\toprule
Variant & $\kappa$ & Acc & F1 & $\Delta\kappa$ \\
\midrule
\rowcolor{GroupGray}
\multicolumn{5}{c}{\emph{Component Ablations}} \\
\rowcolor{OursBlue}
\textbf{Ours (full)} & \textbf{.761}$\pm$.006 & .618 & .511 & -- \\
\quad w/o severity MIL & .737$\pm$.009 & .579 & .495 & $-.024$ \\
\quad w/o competitive assignment & .740$\pm$.005 & .586 & .489 & $-.021$ \\
\quad w/ MLP classification head & .715$\pm$.007 & .596 & .498 & $-.046$ \\
\quad w/o degradation consistency & .749$\pm$.004 & .607 & .505 & $-.012$ \\
\quad w/o positional regularization & .756$\pm$.002 & \textbf{.620} & \textbf{.512} & $-.005$ \\
\midrule
\rowcolor{GroupGray}
\multicolumn{5}{c}{\emph{Reference Classifier}} \\
Global feature classifier (ERM) & .710$\pm$.006 & .598 & .499 & $-.051$ \\
\bottomrule
\end{tabularx}
\end{table}

Table~\ref{tab:ablation} evaluates the contributions of our components under the six domain LODO protocol. Removing severity guided MIL and competitive assignment reduces mean QWK from $0.761$ to $0.737$ and $0.740$, respectively, confirming the importance of clinical supervision and competition between lesion and background prototypes. Replacing the 19 parameter additive ordinal head with an MLP reduces QWK to $0.715$, despite identical prototype inputs, classification loss, and head learning rate. This highlights the advantage of explicitly structuring grading decisions around prototype contributions.

Removing degradation consistency and positional regularization reduces QWK to $0.749$ and $0.756$, respectively. Positional regularization provides a smaller QWK gain, while slightly reducing accuracy and macro F1. The global feature classifier (ERM) achieves $0.710$, further supporting evidence based grading. Additional spatial shortcut analyses are provided in the appendix.

\subsection{Prototype Analysis}
\label{sec:prototype_analysis}

We analyze the learned prototypes from three complementary perspectives: whether their spatial responses align with retinal lesions, whether they influence severity predictions, and whether severity guided supervision transfers to other prototype architectures. Lesion annotations are used only for retrospective evaluation and visualization, never for training.

\begin{figure}[t]
\centering
\includegraphics[width=\columnwidth]{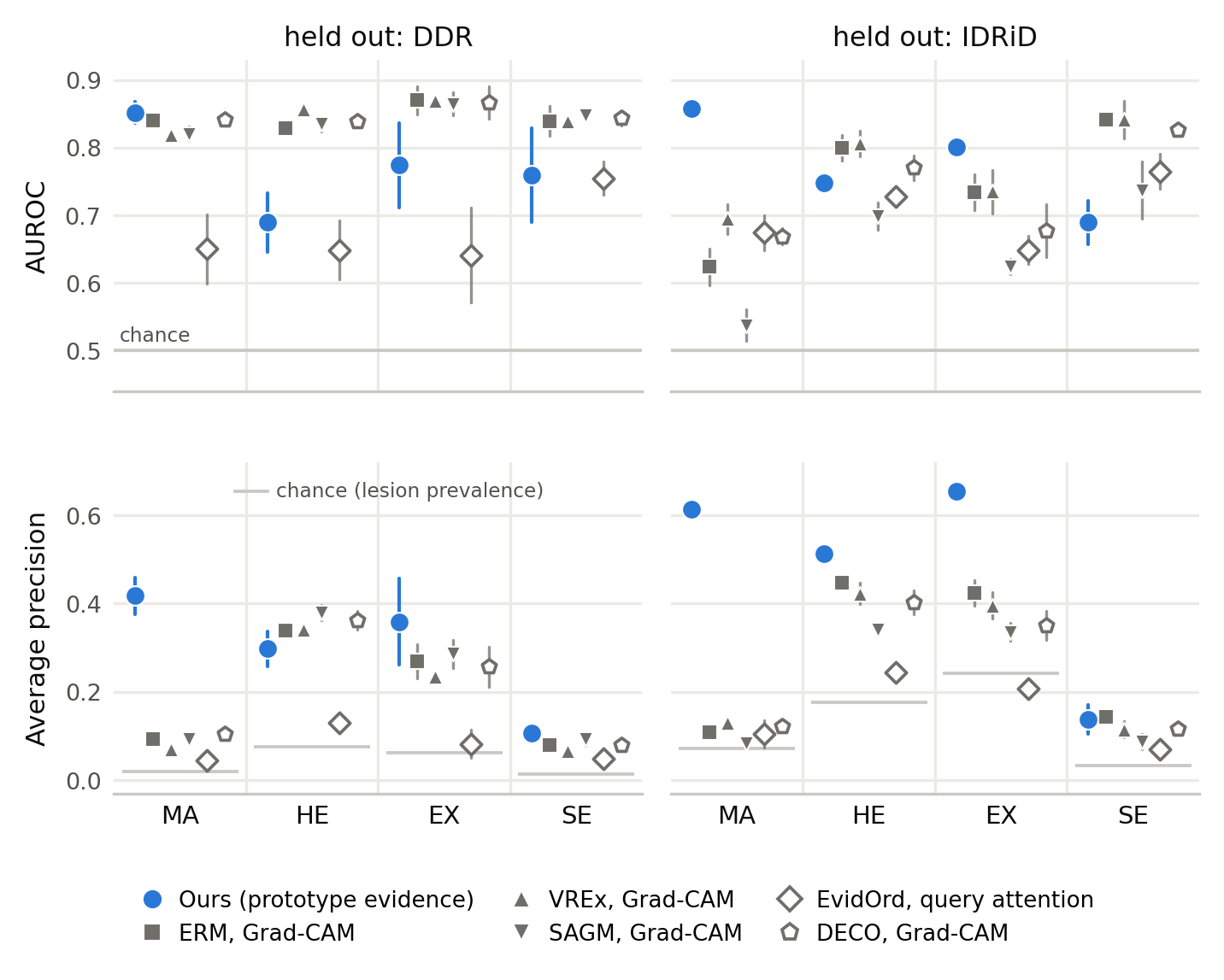}
\caption{Lesion localization on unseen DDR and IDRiD domains. Pixel AUROC (top) and AP (bottom) are reported as mean $\pm$ SD (3 seeds). Gray lines indicate chance levels.}
\label{fig:localization}
\end{figure}

\paragraph{Lesion localization on unseen domains.}\label{sec:localization} We first evaluate the spatial alignment of lesion prototype responses on DDR, IDRiD, and FGADR, which provide pixel annotations for microaneurysms (MA), hemorrhages (HE), hard exudates (EX), and soft exudates (SE). We measure pixel AUROC and average precision (AP), comparing with Grad-CAM~\cite{gradcam} explanations from ERM, VREx, SAGM, and DECO and lesion query attention from EvidOrd. Figure~\ref{fig:localization} shows strong localization on unseen DDR and IDRiD, particularly for small MA, with AP reaching approximately $0.42$ and $0.61$, respectively. FGADR has substantially different annotation granularity for small lesions, complicating direct comparisons of pixel localization scores; its full results and annotation analysis are provided in the appendix.

\begin{figure}[t]
\centering
\includegraphics[width=\columnwidth]{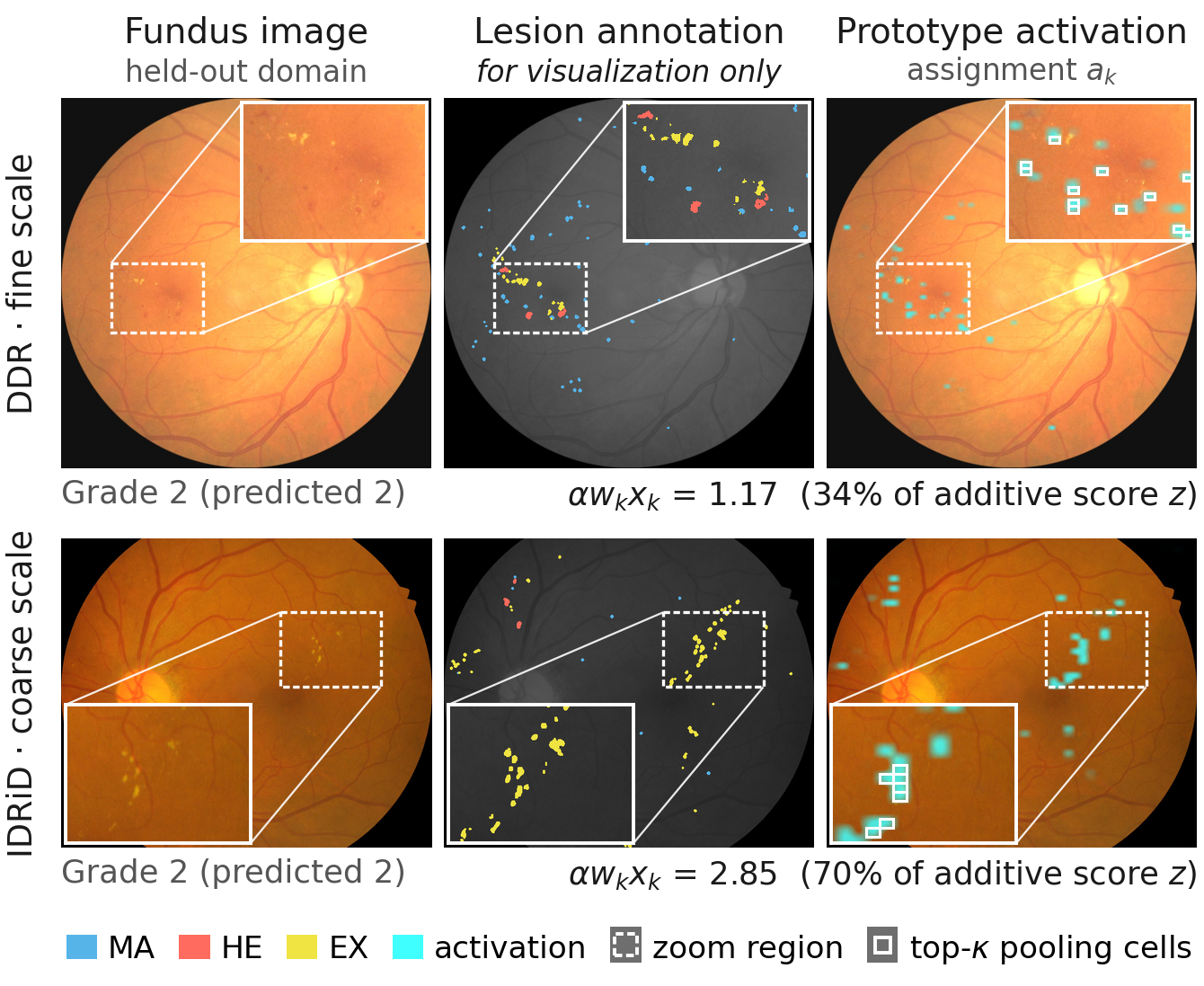}
\caption{Qualitative visualization of lesion prototype evidence on unseen DDR and IDRiD domains. Fine and coarse scale prototypes exhibit spatial responses associated with retinal lesions, contributing $34\%$ and $70\%$ of the additive grading score, respectively. Dashed boxes indicate enlarged regions, while white squares highlight the top-$K$ pooling cells used to compute prototype presence. Lesion annotations are used only for visualization and never during training.}
\label{fig:proto_vis}
\end{figure}

\paragraph{Qualitative lesion evidence.} Figure~\ref{fig:proto_vis} shows representative fine and coarse scale prototypes on unseen domains. The fine scale prototype responds to multiple small pathological regions, while the coarse scale prototype captures clustered exudates. Their respective contributions account for $34\%$ and $70\%$ of the additive grading scores. Together with the lesion annotations shown exclusively for visualization, these examples illustrate how spatial responses associated with pathology enter the grading decision through individual prototype presence scores.

\begin{figure}[t]
\centering
\includegraphics[width=\columnwidth]{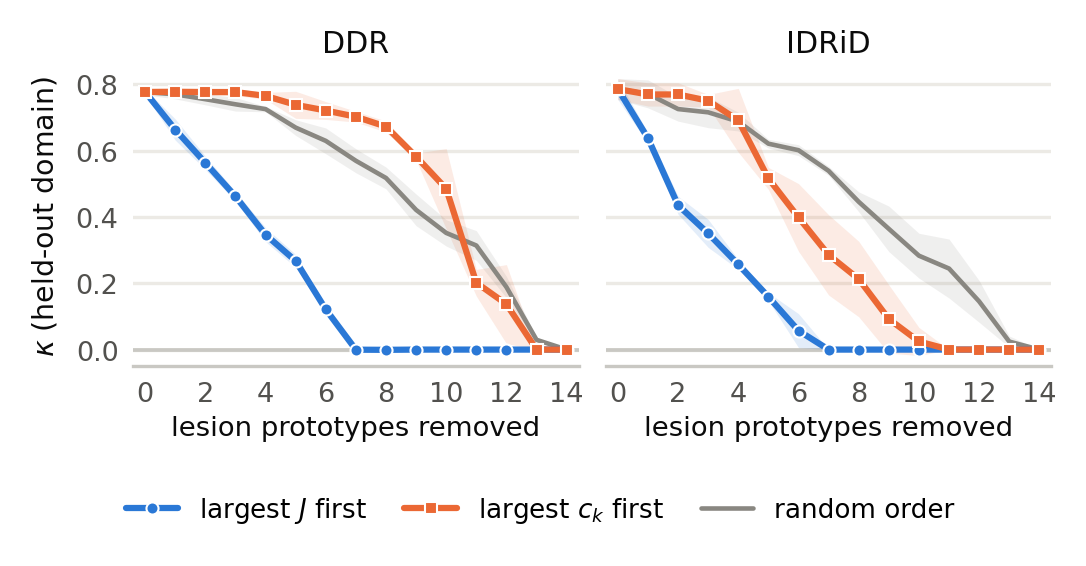}
\caption{Prototype removal on unseen DDR and IDRiD (mean $\pm$ SD, 3 seeds). Removal follows decision contribution, spatial concentration, or random order without retraining. Rankings use retrospective target statistics.}
\label{fig:knockout}
\end{figure}

\begin{figure*}[t]
\centering
\includegraphics[width=\textwidth]{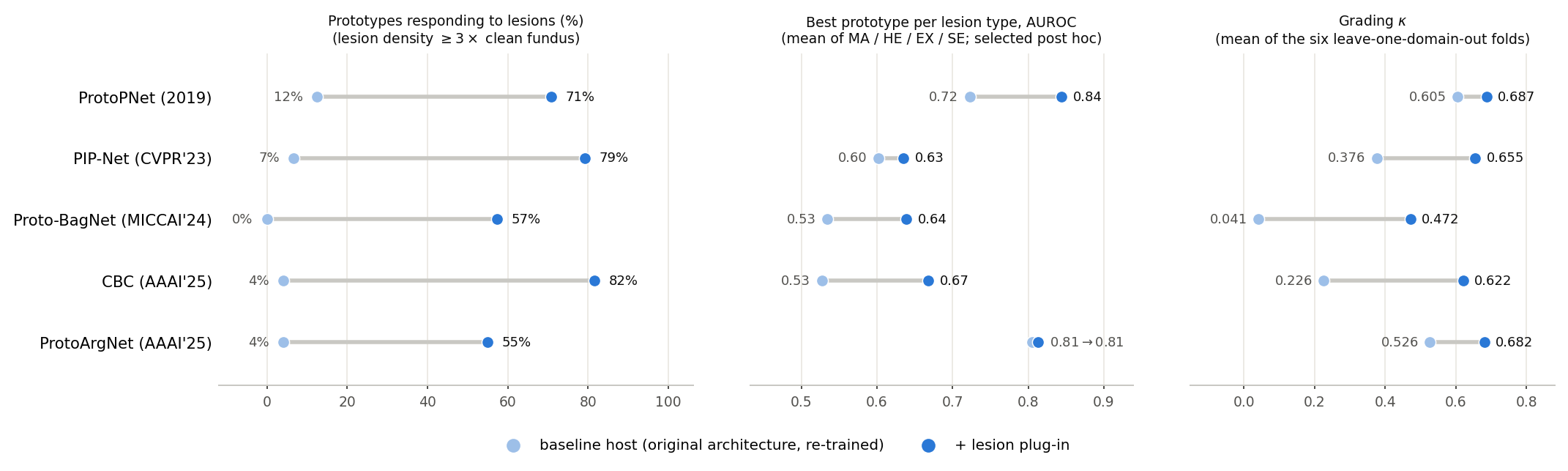}
\caption{Generality of severity guided lesion learning across five prototype architectures. The plug-in improves lesion alignment and grading QWK; the best localization prototype is selected retrospectively using lesion annotations.}
\label{fig:plugin}
\end{figure*}

\paragraph{Prototype contribution to grading.} Spatial alignment alone does not establish that prototype evidence influences severity predictions. We therefore intervene on this evidence by progressively setting prototype presence scores to zero at inference, without retraining or renormalizing the remaining assignments. We compare removal in descending order of decision contribution $J_k=w_k\mathbb{E}[x_k]$, spatial concentration $c_k$, and random order. Rankings are computed retrospectively from all images in each unseen target domain without severity or lesion annotations and are not used to adapt the model. As shown in Fig.~\ref{fig:knockout}, removing the prototypes with the largest decision contributions causes QWK to decline substantially faster than random removal on both DDR and IDRiD, reaching zero after seven of the fourteen lesion prototypes are removed. By contrast, removal based on spatial concentration follows different trends, indicating that persistent coordinate preference is not equivalent to decision importance.

\paragraph{Generality across prototype architectures.}
\label{sec:plugin} 
We evaluate the generality of severity guided lesion supervision by integrating it into five prototype architectures: ProtoPNet~\cite{protopnet}, PIP-Net~\cite{pipnet}, Proto-BagNet~\cite{protobagnet}, CBC~\cite{cbc}, and ProtoArgNet~\cite{protoargnet}. Each architecture is retrained with and without our supervision under the six domain LODO protocol using one seed, with prototype analyses conducted on three annotated domains. As shown in Fig.~\ref{fig:plugin}, mean grading QWK improves from $0.355$ to $0.624$, with gains across all five architectures. The proportion of lesion responsive prototypes increases from $5.4\%$ to $68.8\%$, defined by activation density within annotated lesions at least three times that in clean fundus regions. Best prototype localization AUROC improves in four architectures and remains comparable in ProtoArgNet, with lesion annotations used only for retrospective evaluation. These results demonstrate the broader applicability of our supervision strategy to both pathological alignment and domain generalized grading.

\subsection{Robustness to Image Degradations}
\label{sec:degradation}

\begin{table}[t]
\centering
\caption{Robustness to image degradations on unseen DDR, IDRiD, and FGADR domains. Clean is QWK on original images; other columns report changes relative to Clean. Results are averaged over three seeds and three domains.}
\label{tab:degrade}
\small
\setlength{\tabcolsep}{3pt}
\renewcommand{\arraystretch}{1.12}
\begin{tabular}{@{}lcccc@{}}
\toprule
Method & Clean & Blur & Half res. & Noise \\
\midrule
ERM & .648 & -.053 & -.059 & -.065 \\
VREx~\cite{vrex} & .673 & -.053 & -.054 & -.068 \\
SAGM~\cite{sagm} & .660 & -.049 & -.039 & -.047 \\
MixStyle~\cite{mixstyle} & .657 & -.020 & -.005 & -.050 \\
DECO~\cite{deco} & .656 & -.038 & -.019 & -.072 \\
AdvBlur~\cite{advblur} & .646 & -.063 & -.074 & -.050 \\
EvidOrd~\cite{evidord} & .650 & -.099 & -.093 & -.089 \\
Ours w/o Deg. Cons. & .690 & -.081 & -.082 & -.056 \\
\midrule
\rowcolor{OursBlue}
\textbf{Ours} & \textbf{.704} & \textbf{-.004} & \textbf{-.003} & \textbf{-.036} \\
\bottomrule
\end{tabular}
\end{table}

We assess robustness to blur, reduced resolution, and sensor noise on the unseen DDR, IDRiD, and FGADR domains. Table~\ref{tab:degrade} reports clean QWK and the changes caused by each perturbation. Our method attains the highest clean QWK ($0.704$) and the smallest loss under every perturbation: $0.004$ for blur, $0.003$ for reduced resolution, and $0.036$ for sensor noise. By comparison, ERM loses $0.053$, $0.059$, and $0.065$, respectively.

A direct ablation clarifies the role of lesion evidence preservation. Without degradation consistency, QWK decreases by $0.081$ under blur and $0.082$ under reduced resolution, compared with only $0.004$ and $0.003$ for the full model. The corresponding clean QWK values are $0.690$ and $0.704$. These findings support the role of prototype guided consistency in maintaining stable grading decisions when subtle pathological details are degraded.

\section{Conclusion}
\label{sec:conclusion}

We presented LWM, a lesion grounded framework for domain generalized DR grading that constructs predictions directly from pathological evidence learned using severity labels alone. Competitive prototype learning and severity guided MIL encourage lesion associated representations, while additive ordinal scoring makes their contributions to grading explicit. Evidence preservation and spatial regularization further support generalization. Experiments across six datasets demonstrate strong grading performance, with localization and prototype removal analyses validating pathological alignment and decision relevance. Additional evaluations establish robustness to image degradations and the transferability of our supervision strategy across five prototype architectures. These findings support pathology grounded decision making as an effective direction for generalizable DR grading.

\section{LLM Usage Statement}

In this work, we used ChatGPT (OpenAI) and Claude (Anthropic) to assist with manuscript polishing, literature searches, and code optimization and debugging. We did not use generative AI tools for the core research methodology, experimental design, or interpretation of results. We have reviewed all AI-assisted work. All AI-assisted code was manually reviewed and verified by the authors. We take responsibility for the final
content of this work, including text, claims, or artifacts produced with the aid of generative AI.

{\small
\bibliographystyle{ieeenat_fullname}
\bibliography{main}

@String(CVPR= {IEEE Conf. Comput. Vis. Pattern Recog.})

@String(ICCV= {Int. Conf. Comput. Vis.})

@String(ICASSP=	{ICASSP})

@String(AAAI = {AAAI})

@String(CVPR  = {CVPR})

@String(ICCV  = {ICCV})

@Article{idrid,
AUTHOR = {Porwal, Prasanna and Pachade, Samiksha and Kamble, Ravi and Kokare, Manesh and Deshmukh, Girish and Sahasrabuddhe, Vivek and Meriaudeau, Fabrice},
TITLE = {Indian Diabetic Retinopathy Image Dataset ({IDRiD}): A Database for Diabetic Retinopathy Screening Research},
JOURNAL = {Data},
VOLUME = {3},
YEAR = {2018},
NUMBER = {3},
ISSN = {2306-5729},
DOI = {10.3390/data3030025},
  note = {Art. no. 25},
}

@article{messidor2,
  author = {Decencière, Etienne and Zhang, Xiwei and Cazuguel, Guy and Lay, Bruno and Cochener, Béatrice and Trone, Caroline and Gain, Philippe and Ordonez, Richard and Massin, Pascale and Erginay, Ali and Charton, Béatrice and Klein, Jean-Claude},
  year = {2014},
  month = {Aug.},
  pages = {231--234},
  title = {Feedback on a Publicly Distributed Image Database: The {Messidor} Database},
  volume = {33},
  journal = {Image Analysis \& Stereology},
  doi = {10.5566/ias.1155},
  number = {3},
}

@article{messidor1,
  title={Automated analysis of retinal images for detection of referable diabetic retinopathy},
  author={Abr{\`a}moff, Michael D and Folk, James C and Han, Dennis P and Walker, Jonathan D and Williams, David F and Russell, Stephen R and Massin, Pascale and Cochener, Beatrice and Gain, Philippe and Tang, Li and others},
  journal={JAMA Ophthalmology},
  volume={131},
  number={3},
  pages={351--357},
  year={2013}
}

@misc{aptos,
    author = {Karthik and Maggie and Sohier Dane},
    title = {APTOS 2019 Blindness Detection},
    year = {2019},
    howpublished = {\url{https://www.kaggle.com/competitions/aptos2019-blindness-detection}},
    note = {Kaggle}
}

@article{ddr,
  title={Diagnostic assessment of deep learning algorithms for diabetic retinopathy screening},
  author={Li, Tao and Gao, Yingqi and Wang, Kai and Guo, Song and Liu, Hanruo and Kang, Hong},
  journal={Information Sciences},
  volume={501},
  pages={511--522},
  year={2019}
}

@article{fgadr,
  title={A benchmark for studying diabetic retinopathy: segmentation, grading, and transferability},
  author={Zhou, Yi and Wang, Boyang and Huang, Lei and Cui, Shanshan and Shao, Ling},
  journal={IEEE Transactions on Medical Imaging},
  volume={40},
  number={3},
  pages={818--828},
  year={2021},
  doi = {10.1109/TMI.2020.3037771}
}

@article{deepdrid,
  title={Deepdrid: Diabetic retinopathy—grading and image quality estimation challenge},
  author={Liu, Ruhan and Wang, Xiangning and Wu, Qiang and Dai, Ling and Fang, Xi and Yan, Tao and Son, Jaemin and Tang, Shiqi and Li, Jiang and Gao, Zijian and others},
  journal={Patterns},
  volume={3},
  number={6},
  year={2022},
  publisher={Elsevier}
}

@InProceedings{gradcam,
author = {Selvaraju, Ramprasaath R. and Cogswell, Michael and Das, Abhishek and Vedantam, Ramakrishna and Parikh, Devi and Batra, Dhruv},
title = {Grad-CAM: Visual Explanations From Deep Networks via Gradient-Based Localization},
booktitle = {Proceedings of the IEEE International Conference on Computer Vision (ICCV)},
month = {Oct},
year = {2017}
}

@inproceedings{mil,
  title={Attention-based deep multiple instance learning},
  author={Ilse, Maximilian and Tomczak, Jakub and Welling, Max},
  booktitle={International Conference on Machine Learning},
  pages={2127--2136},
  year={2018},
  organization={PMLR}
}

@article{icdr,
  title={Proposed international clinical diabetic retinopathy and diabetic macular edema disease severity scales},
  author={Wilkinson, Charles P and Ferris III, Frederick L and Klein, Ronald E and Lee, Paul P and Agardh, Carl David and Davis, Matthew and Dills, Diana and Kampik, Anselm and Pararajasegaram, Rangasamy and Verdaguer, Juan T and others},
  journal={Ophthalmology},
  volume={110},
  number={9},
  pages={1677--1682},
  year={2003},
  publisher={Elsevier}
}

@inproceedings{lat,
  title={Lesion-aware transformers for diabetic retinopathy grading},
  author={Sun, Rui and Li, Yihao and Zhang, Tianzhu and Mao, Zhendong and Wu, Feng and Zhang, Yongdong},
  booktitle={2021 IEEE/CVF Conference on Computer Vision and Pattern Recognition (CVPR)},
  pages={10933--10942},
  year={2021},
  organization={IEEE}
}

@inproceedings{weaklesion,
  title={Diabetic retinopathy grading with weakly-supervised lesion priors},
  author={Hou, Junlin and Xiao, Fan and Xu, Jilan and Feng, Rui and Zhang, Yuejie and Zou, Haidong and Lu, Lina and Xue, Wenwen},
  booktitle={ICASSP 2023-2023 IEEE International Conference on Acoustics, Speech and Signal Processing (ICASSP)},
  pages={1--5},
  year={2023},
  organization={IEEE}
}

@inproceedings{swin,
  title={Swin transformer: Hierarchical vision transformer using shifted windows},
  author={Liu, Ze and Lin, Yutong and Cao, Yue and Hu, Han and Wei, Yixuan and Zhang, Zheng and Lin, Stephen and Guo, Baining},
  booktitle={2021 IEEE/CVF International Conference on Computer Vision (ICCV)},
  pages={9992--10002},
  year={2021},
  organization={IEEE}
}

@inproceedings{mixstyle,
title={Domain Generalization with MixStyle},
author={Kaiyang Zhou and Yongxin Yang and Yu Qiao and Tao Xiang},
booktitle={International Conference on Learning Representations},
year={2021}
}

@article{meanteacher,
  title={Mean teachers are better role models: Weight-averaged consistency targets improve semi-supervised deep learning results},
  author={Tarvainen, Antti and Valpola, Harri},
  journal={Advances in Neural Information Processing Systems},
  volume={30},
  year={2017}
}

@article{swad,
  title={Swad: Domain generalization by seeking flat minima},
  author={Cha, Junbum and Chun, Sanghyuk and Lee, Kyungjae and Cho, Han-Cheol and Park, Seunghyun and Lee, Yunsung and Park, Sungrae},
  journal={Advances in Neural Information Processing Systems},
  volume={34},
  pages={22405--22418},
  year={2021}
}

@inproceedings{vrex,
  title={Out-of-distribution generalization via risk extrapolation (rex)},
  author={Krueger, David and Caballero, Ethan and Jacobsen, Joern-Henrik and Zhang, Amy and Binas, Jonathan and Zhang, Dinghuai and Le Priol, Remi and Courville, Aaron},
  booktitle={International Conference on Machine Learning},
  pages={5815--5826},
  year={2021},
  organization={PMLR}
}

@inproceedings{fishr,
  title={Fishr: Invariant gradient variances for out-of-distribution generalization},
  author={Rame, Alexandre and Dancette, Corentin and Cord, Matthieu},
  booktitle={International Conference on Machine Learning},
  pages={18347--18377},
  year={2022},
  organization={PMLR}
}

@inproceedings{sagm,
  title={Sharpness-aware gradient matching for domain generalization},
  author={Wang, Pengfei and Zhang, Zhaoxiang and Lei, Zhen and Zhang, Lei},
  booktitle={2023 IEEE/CVF Conference on Computer Vision and Pattern Recognition (CVPR)},
  pages={3769--3778},
  year={2023},
  organization={IEEE}
}

@inproceedings{ermpp,
  title={Erm++: An improved baseline for domain generalization},
  author={Teterwak, Piotr and Saito, Kuniaki and Tsiligkaridis, Theodoros and Saenko, Kate and Plummer, Bryan},
  booktitle={Proceedings of the Winter Conference on Applications of Computer Vision},
  pages={8514--8524},
  year={2025}
}

@inproceedings{drgen,
  title={Drgen: Domain generalization in diabetic retinopathy classification},
  author={Atwany, Mohammad and Yaqub, Mohammad},
  booktitle={International Conference on Medical Image Computing and Computer-Assisted Intervention},
  pages={635--644},
  year={2022},
  organization={Springer}
}

@inproceedings{gdrnet,
  title={Towards generalizable diabetic retinopathy grading in unseen domains},
  author={Che, Haoxuan and Cheng, Yuhan and Jin, Haibo and Chen, Hao},
  booktitle={International Conference on Medical Image Computing and Computer-Assisted Intervention},
  pages={430--440},
  year={2023},
  organization={Springer}
}

@inproceedings{dg,
title={In Search of Lost Domain Generalization},
author={Ishaan Gulrajani and David Lopez-Paz},
booktitle={International Conference on Learning Representations},
year={2021},
url={https://openreview.net/forum?id=lQdXeXDoWtI}
}

@inproceedings{deco,
  title={Generalizing to unseen domains in diabetic retinopathy with disentangled representations},
  author={Xia, Peng and Hu, Ming and Tang, Feilong and Li, Wenxue and Zheng, Wenhao and Ju, Lie and Duan, Peibo and Yao, Huaxiu and Ge, Zongyuan},
  booktitle={International Conference on Medical Image Computing and Computer-Assisted Intervention},
  pages={427--437},
  year={2024},
  organization={Springer}
}

@article{gad,
  title={GAD: Domain generalized diabetic retinopathy grading by grade-aware de-stylization},
  author={Bi, Qi and Yi, Jingjun and Zheng, Hao and Zhan, Haolan and Huang, Yawen and Ji, Wei and Li, Yuexiang and Zheng, Yefeng},
  journal={Pattern Recognition},
  volume={164},
  pages={111484},
  year={2025},
  publisher={Elsevier}
}

@inproceedings{loasp,
  title={Low-rank adaptive structural priors for generalizable diabetic retinopathy grading},
  author={Wang, Yunxuan and Yin, Rui and Tan, Yumei and Chen, Hao and Xia, Haiying},
  booktitle={2025 International Joint Conference on Neural Networks (IJCNN)},
  pages={1--8},
  year={2025},
  organization={IEEE}
}

@article{advblur,
  title={AdvBlur: Adversarial Blur for Robust Diabetic Retinopathy Classification and Cross-Domain Generalization},
  author={Kanagalingam, Heethanjan and Pathmanathan, Thenukan and Vathanakumar, Mokeeshan and Mukunthan, Tharmakulasingam},
  journal={arXiv preprint arXiv:2510.24000},
  year={2025}
}

@article{evidord,
  title={Uncertainty-Aware Ordinal Deep Learning for cross-Dataset Diabetic Retinopathy Grading},
  author={Bellaj, Ali El and Benradi, Aya and Youssoufi, Salman El and Marzouki, Taha El and Cheddadi, Mohammed-Amine},
  journal={arXiv preprint arXiv:2602.10315},
  year={2026}
}

@article{protopnet,
  title={This looks like that: deep learning for interpretable image recognition},
  author={Chen, Chaofan and Li, Oscar and Tao, Daniel and Barnett, Alina and Rudin, Cynthia and Su, Jonathan K},
  journal={Advances in Neural Information Processing Systems},
  volume={32},
  year={2019}
}

@inproceedings{pipnet,
  title={Pip-net: Patch-based intuitive prototypes for interpretable image classification},
  author={Nauta, Meike and Schl{\"o}tterer, J{\"o}rg and Van Keulen, Maurice and Seifert, Christin},
  booktitle={2023 IEEE/CVF Conference on Computer Vision and Pattern Recognition (CVPR)},
  pages={2744--2753},
  year={2023},
  organization={IEEE}
}

@inproceedings{protobagnet,
  title={This actually looks like that: Proto-BagNets for local and global interpretability-by-design},
  author={Djoumessi, Kerol and Bah, Bubacarr and K{\"u}hlewein, Laura and Berens, Philipp and Koch, Lisa},
  booktitle={International Conference on Medical Image Computing and Computer-Assisted Intervention},
  pages={718--728},
  year={2024},
  organization={Springer}
}

@inproceedings{cbc,
  title={A robust prototype-based network with interpretable RBF classifier foundations},
  author={Saralajew, Sascha and Rana, Ashish and Villmann, Thomas and Shaker, Ammar},
  booktitle={Proceedings of the AAAI Conference on Artificial Intelligence},
  volume={39},
  number={19},
  pages={20273--20282},
  year={2025}
}

@article{ordinal,
author = {Wenzhi Cao and Vahid Mirjalili and Sebastian Raschka},
title = {Rank consistent ordinal regression for neural networks with application to age estimation},
journal = {Pattern Recognition Letters},
volume = {140},
pages = {325-331},
year = {2020},
issn = {0167-8655},
}

@inproceedings{protoargnet,
  title={ProtoArgNet: Interpretable image classification with super-prototypes and argumentation},
  author={Ayoobi, Hamed and Potyka, Nico and Toni, Francesca},
  booktitle={Proceedings of the AAAI Conference on Artificial Intelligence},
  volume={39},
  number={2},
  pages={1791--1799},
  year={2025}
}

@InProceedings{resnet,
author = {He, Kaiming and Zhang, Xiangyu and Ren, Shaoqing and Sun, Jian},
title = {Deep Residual Learning for Image Recognition},
booktitle = {Proceedings of the IEEE Conference on Computer Vision and Pattern Recognition (CVPR)},
month = {June},
year = {2016}
}

@article{diabetic,
  title={Diabetic retinopathy},
  author={Fong, Donald S and Aiello, Lloyd P and Ferris III, Frederick L and Klein, Ronald},
  journal={Diabetes Care},
  volume={27},
  number={10},
  year={2004},
  pages = {2540--2553},
  doi = {10.2337/diacare.27.10.2540},
}
}

\clearpage
\appendix
\setcounter{table}{0}
\setcounter{figure}{0}
\renewcommand{\thetable}{S\arabic{table}}
\renewcommand{\thefigure}{S\arabic{figure}}

\renewcommand{\topfraction}{0.95}
\renewcommand{\dbltopfraction}{0.95}
\renewcommand{\textfraction}{0.04}
\renewcommand{\floatpagefraction}{0.75}
\renewcommand{\dblfloatpagefraction}{0.75}
\setcounter{topnumber}{4}
\setcounter{totalnumber}{6}
\setcounter{dbltopnumber}{2}

\section{Detailed Grading Results Across Unseen Domains}
\label{sec:supp_lodo}

Table~\ref{tab:supp_lodo} expands the main-paper grading comparison into domain-wise QWK, accuracy, and macro-F1 for each method. All values follow the main paper and are averaged over three training seeds. Our method achieves the highest six-domain means of $0.761$ QWK, $0.618$ accuracy, and $0.511$ macro-F1; individual-domain results illustrate differences among ordinal agreement, exact grading accuracy, and class-balanced performance.

\begin{table*}[!t]
\centering
\caption{Detailed LODO grading results across six unseen domains (three training seeds). We report quadratic weighted kappa ($\kappa$), accuracy (Acc), and macro-F1 (F1). Bold and underlined values indicate the highest and second-highest reported values, including ties.}
\label{tab:supp_lodo}
\footnotesize
\setlength{\tabcolsep}{4.3pt}
\renewcommand{\arraystretch}{1.02}
\begin{tabular}{ll*{7}{c}}
\toprule
& & \multicolumn{6}{c}{Held-out Domain} & \multicolumn{1}{c}{Overall} \\
\cmidrule(lr){3-8}
\cmidrule(lr){9-9}
Method & Metric & FGADR & APTOS & DDR & IDRiD & Mess-2 & DeepDRiD & Mean \\
\midrule

\rowcolor{GroupGray}
\multicolumn{9}{l}{\emph{Standard Baselines}} \\
ERM (Swin-T) & $\kappa$ & .462 & .795 & .737 & .745 & .783 & .740 & .710 \\
 & Acc & .467 & .579 & .679 & \underline{.578} & .683 & \underline{.603} & .598 \\
 & F1 & .376 & .455 & .521 & .536 & .597 & \underline{.507} & .499 \\[1pt]
ERM (ResNet-50) & $\kappa$ & .463 & .831 & .726 & .765 & .776 & .693 & .709 \\
 & Acc & .413 & \underline{.697} & \underline{.698} & .571 & .687 & .558 & .604 \\
 & F1 & .328 & .479 & .515 & .515 & .560 & .439 & .473 \\

\addlinespace[2pt]
\rowcolor{GroupGray}
\multicolumn{9}{l}{\emph{General DG Methods}} \\
MixStyle\textsuperscript{$\ast$}~\cite{mixstyle} & $\kappa$ & .485 & .794 & .753 & .734 & .818 & .735 & .720 \\
 & Acc & .475 & .586 & .686 & .569 & .705 & .596 & .603 \\
 & F1 & .389 & .461 & .528 & .515 & \textbf{.641} & .489 & .504 \\[1pt]
SWAD\textsuperscript{$\ast$}~\cite{swad} & $\kappa$ & .478 & .802 & .731 & .765 & .775 & .737 & .715 \\
 & Acc & .480 & .592 & .670 & \underline{.578} & .673 & .596 & .598 \\
 & F1 & .388 & .459 & .514 & .535 & .601 & .502 & .500 \\[1pt]
VREx\textsuperscript{$\ast$}~\cite{vrex} & $\kappa$ & .454 & .823 & \textbf{.779} & \underline{.785} & .792 & .717 & \underline{.725} \\
 & Acc & .467 & .620 & .690 & .577 & .701 & .573 & .605 \\
 & F1 & .364 & .455 & .513 & .528 & .597 & .472 & .488 \\[1pt]
Fishr\textsuperscript{$\ast$}~\cite{fishr} & $\kappa$ & .447 & .795 & .738 & .729 & .794 & .724 & .705 \\
 & Acc & .462 & .578 & .683 & \underline{.578} & .694 & .590 & .597 \\
 & F1 & .361 & .453 & .525 & .539 & \underline{.631} & .497 & .501 \\[1pt]
SAGM\textsuperscript{$\ast$}~\cite{sagm} & $\kappa$ & .515 & .813 & .745 & .720 & .798 & .737 & .721 \\
 & Acc & \textbf{.502} & .626 & .688 & .572 & .702 & .593 & .614 \\
 & F1 & \underline{.412} & .470 & \textbf{.536} & .520 & .625 & .494 & \underline{.509} \\[1pt]
ERM++\textsuperscript{$\ast$}~\cite{ermpp} & $\kappa$ & .464 & .768 & .676 & \textbf{.791} & .662 & .713 & .679 \\
 & Acc & .479 & .499 & .616 & .571 & .626 & .561 & .559 \\
 & F1 & .361 & .415 & .459 & \underline{.542} & .502 & .443 & .454 \\

\addlinespace[2pt]
\rowcolor{GroupGray}
\multicolumn{9}{l}{\emph{DR-specific Methods}} \\
GDRNet\textsuperscript{$\dagger$}~\cite{gdrnet} & $\kappa$ & \underline{.525} & .759 & .737 & .780 & .772 & .717 & .715 \\
 & Acc & .446 & .586 & .685 & .555 & .691 & .577 & .590 \\
 & F1 & .370 & .441 & .494 & .527 & .565 & .447 & .474 \\[1pt]
DRGen\textsuperscript{$\ast$}~\cite{drgen} & $\kappa$ & .471 & .800 & .733 & .749 & .784 & .715 & .709 \\
 & Acc & .475 & .589 & .669 & .568 & .679 & .582 & .594 \\
 & F1 & .379 & .459 & .517 & .523 & .613 & .490 & .497 \\[1pt]
DECO\textsuperscript{$\ddagger$}~\cite{deco} & $\kappa$ & .458 & .843 & .754 & .755 & .801 & .709 & .720 \\
 & Acc & .458 & .691 & .688 & .570 & .692 & .580 & .613 \\
 & F1 & .375 & \underline{.488} & \underline{.535} & .525 & .607 & .474 & .500 \\[1pt]
AdvBlur\textsuperscript{$\ddagger$}~\cite{advblur} & $\kappa$ & .434 & .811 & .746 & .759 & .789 & .708 & .708 \\
 & Acc & .455 & .625 & .682 & .555 & .692 & .579 & .598 \\
 & F1 & .361 & .472 & .518 & .517 & .612 & .492 & .495 \\[1pt]
GAD\textsuperscript{$\ddagger$}~\cite{gad} & $\kappa$ & .466 & .782 & .726 & .759 & .777 & .735 & .708 \\
 & Acc & .470 & .584 & .677 & .577 & .677 & .591 & .596 \\
 & F1 & .375 & .462 & .506 & \textbf{.544} & .613 & .495 & .499 \\[1pt]
LoASP\textsuperscript{$\ddagger$\S}~\cite{loasp} & $\kappa$ & .447 & \textbf{.852} & .757 & .747 & .753 & \underline{.745} & .717 \\
 & Acc & .407 & \textbf{.721} & .691 & \textbf{.601} & .644 & \textbf{.608} & .612 \\
 & F1 & .334 & \textbf{.496} & .513 & .523 & .567 & .488 & .487 \\[1pt]
EvidOrd\textsuperscript{$\ddagger$}~\cite{evidord} & $\kappa$ & .449 & .841 & \underline{.777} & .724 & \underline{.819} & .730 & .723 \\
 & Acc & .436 & .694 & \textbf{.706} & .559 & \underline{.712} & .587 & \underline{.616} \\
 & F1 & .370 & .476 & \textbf{.536} & .499 & .601 & .504 & .498 \\

\midrule
\rowcolor{OursBlue}
\textbf{Ours} & $\kappa$ & \textbf{.551} & \underline{.846} & \underline{.777} & \underline{.785} & \textbf{.849} & \textbf{.755} & \textbf{.761} \\
\rowcolor{OursBlue}
 & Acc & \underline{.495} & .644 & .654 & \textbf{.601} & \textbf{.731} & .584 & \textbf{.618} \\
\rowcolor{OursBlue}
 & F1 & \textbf{.437} & .457 & .481 & .541 & .624 & \textbf{.528} & \textbf{.511} \\
\bottomrule
\end{tabular}
\end{table*}

\noindent\textit{Implementation notes.} $^{\ast}$ Reimplemented in our unified pipeline using publicly available code; $^{\dagger}$ adapted from official code; $^{\ddagger}$ reimplemented from the original paper without public code; $^{\S}$ uses ResNet-50 because LoASP operates on ResNet bottleneck blocks. All other methods use Swin-T unless specified otherwise. ResNet-50 ERM achieves a mean QWK of $0.709\pm0.016$.

\section{Complete Lesion Localization Results}
\label{sec:supp_localization}

\begin{figure*}[!t]
\centering
\includegraphics[width=.97\textwidth]{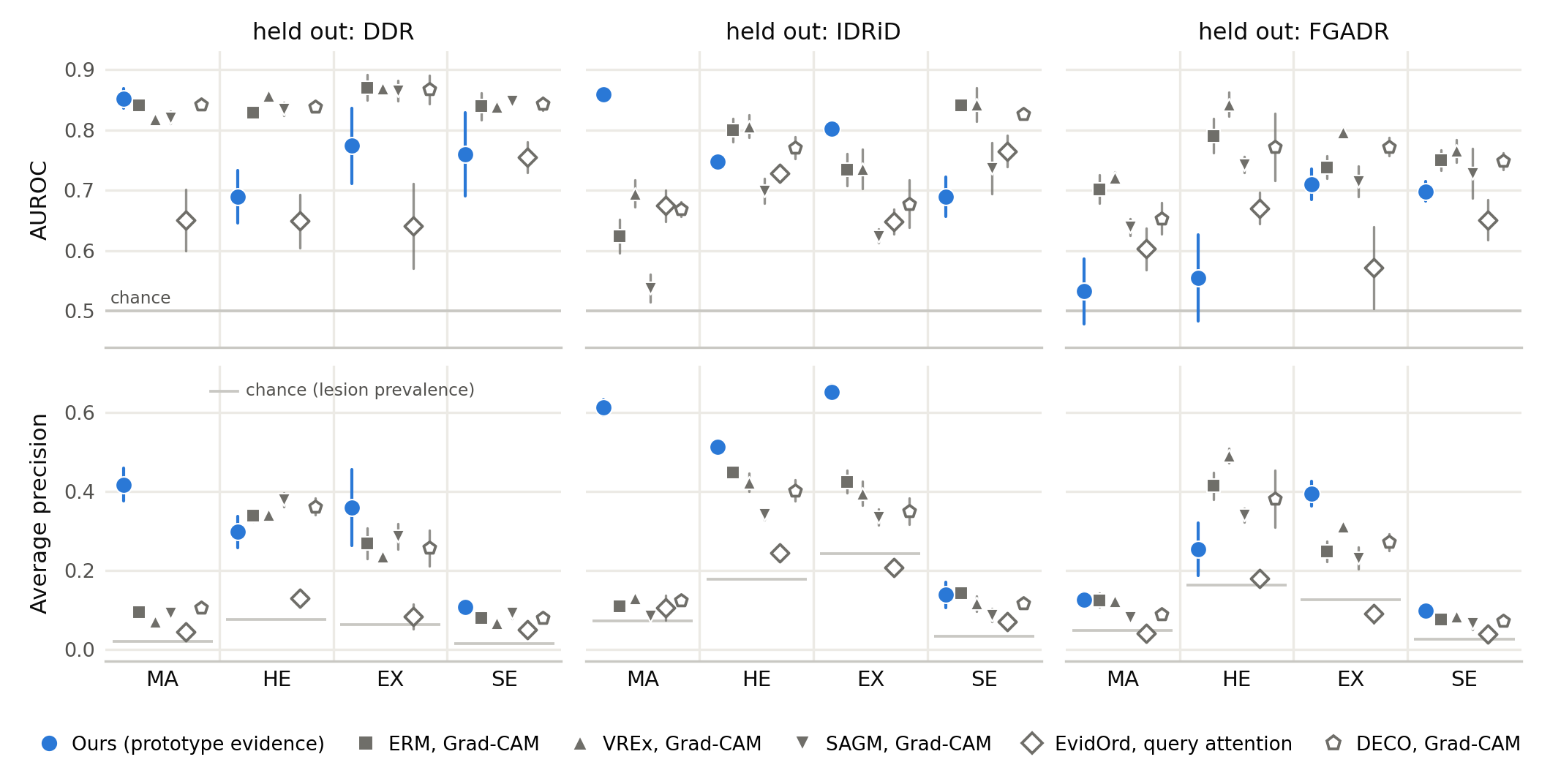}
\caption{Complete lesion localization results on unseen DDR, IDRiD, and FGADR domains. Pixel AUROC (top) and average precision (bottom) are reported as mean $\pm$ SD across three random seeds. Gray lines indicate chance levels (AUROC of 0.5 and lesion prevalence for AP). Lesion annotations are used only for evaluation.}
\label{fig:supp_localization}
\end{figure*}

Figure~\ref{fig:supp_localization} presents the complete lesion localization results on DDR, IDRiD, and FGADR, covering microaneurysms (MA), hemorrhages (HE), hard exudates (EX), and soft exudates (SE). We report pixel AUROC and average precision (AP) across three random seeds. LWM demonstrates strong localization performance on DDR and IDRiD, particularly for MA, while its performance on FGADR is more heterogeneous. Under strict pixel-level evaluation, Grad-CAM baselines outperform LWM on FGADR for several lesion categories, although LWM achieves competitive or stronger precision for certain lesions. These differences motivate our subsequent analyses of annotation granularity, annotation dilation, and spatial tolerance.
 
\section{Annotation Granularity and Localization Evaluation}
\label{sec:supp_annotation}

\paragraph{Annotation granularity across datasets.}
We investigate differences in lesion annotation granularity across DDR, IDRiD, and FGADR. We sample 150 images from each dataset and resize them to a width of 1280 pixels, corresponding to $10\times10$ pixel evaluation cells. Table~\ref{tab:annotation_granularity} reports the median area of annotated connected components and the mean annotation coverage within cells intersecting each lesion category. FGADR exhibits substantially larger annotated regions for microaneurysms (MA) and hard exudates (EX). For example, the median MA component area is 78 pixels squared in FGADR, compared with 12 and 21 in DDR and IDRiD, respectively. The corresponding cell coverage increases from 0.12 and 0.15 to 0.65. Similar differences are observed for EX, whereas hemorrhage (HE) annotations in FGADR have a median component area comparable to IDRiD. These observations reveal substantial differences in annotation granularity across datasets, which may influence pixel-level localization evaluation.

\begin{table}[!htbp]
\centering
\caption{Annotation granularity across three datasets (150 images each). Area denotes median component area (px$^2$), and coverage denotes the mean annotated fraction within intersected $10\times10$ pixel cells.}
\label{tab:annotation_granularity}
\footnotesize
\setlength{\tabcolsep}{4pt}
\renewcommand{\arraystretch}{1.06}
\begin{tabular}{llcccc}
\toprule
Dataset & Measure & MA & HE & EX & SE \\
\midrule
DDR & Area & 12 & 51 & 13 & 217 \\
    & Coverage & .12 & .49 & .34 & .45 \\
\addlinespace[2pt]
IDRiD & Area & 21 & 143 & 19 & 771 \\
      & Coverage & .15 & .51 & .31 & .60 \\
\addlinespace[2pt]
FGADR & Area & 78 & 138 & 73 & 1782 \\
      & Coverage & .65 & .62 & .55 & .70 \\
\bottomrule
\end{tabular}
\end{table}

\paragraph{Sensitivity to annotation dilation.}
To examine how annotation extent affects localization metrics, we artificially dilate the ground-truth lesion masks by a radius $\rho\in\{0,3,6,12\}$ pixels while keeping the predicted evidence maps unchanged. Table~\ref{tab:annotation_dilation} reports MA localization AUROC on DDR and IDRiD. As the annotations expand, LWM exhibits a substantial reduction in AUROC, from 0.852 to 0.641 on DDR and from 0.859 to 0.672 on IDRiD. In contrast, ERM Grad-CAM remains relatively stable, decreasing from 0.840 to 0.826 and from 0.624 to 0.604, respectively. Similar trends are observed for the other lesion categories, although the reductions for LWM are smaller. This difference is consistent with the spatial characteristics of the evidence maps: prototype responses tend to concentrate around localized structures, whereas Grad-CAM produces broader activation regions. Enlarging the annotations introduces additional positive locations beyond the concentrated prototype responses, making their localization scores more sensitive to annotation extent.

\begin{table}[!htbp]
\centering
\caption{Effect of annotation dilation on MA localization AUROC. Ground-truth masks are dilated by $\rho$ pixels while evidence maps remain unchanged.}
\label{tab:annotation_dilation}
\footnotesize
\setlength{\tabcolsep}{4pt}
\renewcommand{\arraystretch}{1.04}
\begin{tabular}{llcccc}
\toprule
Domain & Method & $\rho=0$ & $\rho=3$ & $\rho=6$ & $\rho=12$ \\
\midrule
DDR & Ours & .852 & .773 & .715 & .641 \\
 & ERM (Grad-CAM) & .840 & .837 & .834 & .826 \\
\midrule
IDRiD & Ours & .859 & .799 & .747 & .672 \\
 & ERM (Grad-CAM) & .624 & .613 & .611 & .604 \\
\bottomrule
\end{tabular}
\end{table}

\paragraph{Localization with spatial tolerance.}
We further evaluate lesion localization under a spatial tolerance of $r$ evaluation cells, where $r=0$ corresponds to the standard evaluation in the main paper. For an evidence map $E$ on the $128\times128$ evaluation grid, we apply max pooling with a $(2r+1)\times(2r+1)$ kernel and unit stride, preserving the original spatial resolution. The resulting score at location $i$ is $\widetilde{E}_r(i)=\max_{\|j-i\|_\infty\leq r}E(j)$. Ground-truth annotations remain unchanged, and positive samples are the cells intersecting the corresponding lesion annotations. Negative samples are restricted to clean retinal cells containing no annotated lesions of any category, with at least 90\% fundus coverage and a Chebyshev distance greater than $r$ from every annotated lesion cell. Cells within this exclusion band are omitted to avoid penalizing evidence propagated to neighboring locations by max pooling. Following the main evaluation protocol, we randomly sample up to 1,500 negative cells per image using a fixed random seed, share the sampled negatives across lesion categories, and compute AUROC by pooling evaluated cells across all images in each dataset. The same procedure is applied to LWM and all Grad-CAM baselines.

\begin{table}[!htbp]
\centering
\caption{FGADR localization AUROC under spatial tolerance $r$. Evidence is max pooled within an $r$-cell neighborhood, with nearby negatives excluded. The standard protocol corresponds to $r=0$.}
\label{tab:localization_tolerance}
\footnotesize
\setlength{\tabcolsep}{3.5pt}
\renewcommand{\arraystretch}{1.06}
\begin{tabular}{lccccc}
\toprule
Method & $r$ & MA & HE & EX & SE \\
\midrule
Ours & 0 & .532 & .554 & .710 & .698 \\
     & 4 & .723 & .742 & .911 & .832 \\
\addlinespace[2pt]
ERM (Grad-CAM) & 0 & .702 & .790 & .738 & .750 \\
               & 4 & .695 & .789 & .742 & .730 \\
\addlinespace[2pt]
VREx (Grad-CAM) & 0 & .721 & .842 & .796 & .765 \\
                & 4 & .681 & .805 & .771 & .717 \\
\bottomrule
\end{tabular}
\end{table}

As shown in Table~\ref{tab:localization_tolerance}, spatial tolerance substantially improves LWM's measured localization AUROC on FGADR. At $r=4$, MA, HE, EX, and SE AUROC increase from 0.532, 0.554, 0.710, and 0.698 to 0.723, 0.742, 0.911, and 0.832, respectively. LWM exceeds both Grad-CAM baselines on MA, EX, and SE under this evaluation, while HE remains below both baselines. In contrast, the Grad-CAM methods exhibit relatively small changes or moderate decreases. Similar improvements are observed on DDR and IDRiD, where LWM's MA AUROC increases from 0.852 to 0.903 and from 0.859 to 0.925, respectively.

\paragraph{Discussion.}
These analyses indicate that annotation granularity can substantially affect the measured localization performance of spatially concentrated prototype evidence. Annotation dilation reduces LWM's measured AUROC more strongly than that of Grad-CAM, while spatial tolerance reveals lesion-associated responses that may be underestimated by strict pixel-level evaluation. Importantly, the tolerance protocol changes both evidence aggregation and negative-sample selection, so the resulting gains should not be attributed to max pooling alone. Moreover, annotation differences do not fully explain all localization discrepancies, particularly for HE on FGADR. Overall, the results suggest that evaluation protocols should account for differences in the spatial extent of learned evidence and lesion annotations when assessing weakly supervised localization across datasets.

\section{Generality Across Prototype Architectures}
\label{sec:supp_plugin}

To further evaluate the transferability of severity-guided lesion supervision, we report detailed results for five prototype architectures in Table~\ref{tab:supp_plugin}. Each architecture is retrained under our unified six-domain LODO protocol, both with and without the proposed lesion learning plug-in. We evaluate the proportion of lesion-responsive prototypes, localization AUROC using the aggregated lesion evidence map, best individual prototype localization AUROC, and grading QWK. Each configuration uses one training seed, with grading performance averaged over six held-out domains and localization evaluated on the three datasets with lesion annotations.

\begin{table}[!htbp]
\centering
\caption{Generality of severity-guided lesion learning across five prototype architectures. B: original architecture retrained in our pipeline; A: with the lesion plug-in. $K$ denotes prototype count. Mass and Best denote lesion localization AUROC from the aggregated evidence map and the retrospectively selected best prototype, respectively. One training seed per configuration.}
\label{tab:supp_plugin}
\scriptsize
\setlength{\tabcolsep}{3pt}
\renewcommand{\arraystretch}{1.10}
\begin{tabular}{lccrrrr}
\toprule
Architecture & $K$ & Run & Lesion (\%) & Mass & Best & QWK \\
\midrule
ProtoPNet & 40 & B & 12.5 & .494 & .722 & .605 \\
 & & A & 70.8 & .804 & .844 & .687 \\
\addlinespace[2pt]
PIP-Net & 40 & B & 6.7 & .557 & .602 & .376 \\
 & & A & 79.2 & .757 & .635 & .655 \\
\addlinespace[2pt]
Proto-BagNet & 50 & B & 0.0 & .501 & .535 & .041 \\
 & & A & 57.3 & .582 & .638 & .472 \\
\addlinespace[2pt]
CBC & 40 & B & 4.2 & .517 & .527 & .226 \\
 & & A & 81.7 & .718 & .668 & .622 \\
\addlinespace[2pt]
ProtoArgNet & 40 & B & 4.2 & .524 & .805 & .526 \\
 & & A & 55.0 & .648 & .813 & .682 \\
\bottomrule
\end{tabular}
\end{table}

\paragraph{Results and analysis.}
The proposed plug-in improves grading QWK across all five architectures, increasing the mean from $0.355$ to $0.624$. The proportion of lesion-responsive prototypes also increases substantially across all architectures. Importantly, localization AUROC computed from the aggregated lesion evidence map improves consistently, with the mean increasing from $0.519$ to $0.702$. This suggests that the plug-in improves the overall spatial alignment of prototype evidence with retinal pathology rather than merely enhancing a retrospectively selected individual prototype. Best-prototype AUROC also increases across all five architectures, from $0.638$ to $0.720$ on average, with the smallest improvement observed for ProtoArgNet ($0.805$ to $0.813$). Together, these results support the transferability of severity-guided lesion supervision across diverse prototype architectures beyond the specific design of LWM.

\section{Controlled Spatial Shortcut Analysis}
\label{sec:supp_shortcut}

\paragraph{Experimental setup.}
We investigate shortcut reliance using synthetic visual markers correlated with DR severity during source training. We consider two settings: a corner marker outside the retinal field of view (FOV) and a marker at a fixed location inside the retinal FOV. For each setting, we evaluate on unseen APTOS and DDR domains under three conditions: Clean, without the marker; Correlated, with the training-consistent marker-label relationship; and Reversed, with the opposite relationship. We report quadratic weighted kappa (QWK) and define $\Delta\kappa=\kappa_{\mathrm{Rev}}-\kappa_{\mathrm{Corr}}$. All results are obtained from three training seeds. Both LWM variants use largest-connected-component (LCC) filtering for prototype pooling, consistent with the inference procedure described in the main paper.

\begin{table*}[!t]
\centering
\caption{Grading performance under synthetic spatial shortcuts. Clean, Corr., and Rev. denote marker-free, training-consistent, and correlation-reversed evaluation, respectively. QWK values are averaged across three seeds, and $\Delta\kappa=\mathrm{Rev}-\mathrm{Corr}$ is reported as mean $\pm$ SD. Both LWM variants use LCC-based prototype pooling; ERM and DECO use their respective standard inference procedures.}
\label{tab:supp_shortcut_qwk}
\footnotesize
\setlength{\tabcolsep}{4pt}
\renewcommand{\arraystretch}{1.08}
\begin{tabular}{lcccccccc}
\toprule
& \multicolumn{4}{c}{APTOS} & \multicolumn{4}{c}{DDR} \\
\cmidrule(lr){2-5}
\cmidrule(lr){6-9}
Method & Clean & Corr. & Rev. & $\Delta\kappa$ & Clean & Corr. & Rev. & $\Delta\kappa$ \\
\midrule
\multicolumn{9}{l}{\emph{Outside-FOV marker}} \\
ERM & .815 & .796 & .604 & $-.192\pm.013$ & .680 & .802 & .644 & $-.157\pm.005$ \\
DECO & .860 & .839 & .629 & $-.210\pm.022$ & .658 & .815 & .595 & $-.220\pm.028$ \\
w/o Pos. Reg. & .868 & .869 & .868 & $-.000\pm.000$ & .732 & .733 & .732 & $-.001\pm.000$ \\
Ours & .851 & .851 & .851 & $-.000\pm.001$ & .732 & .732 & .732 & $-.001\pm.000$ \\
\addlinespace[4pt]
\multicolumn{9}{l}{\emph{Inside-FOV marker}} \\
ERM & .835 & .818 & .639 & $-.179\pm.016$ & .681 & .800 & .633 & $-.167\pm.017$ \\
DECO & .858 & .836 & .605 & $-.231\pm.015$ & .672 & .812 & .583 & $-.230\pm.022$ \\
w/o Pos. Reg. & .862 & .838 & .564 & $-.274\pm.032$ & .698 & .832 & .504 & $-.329\pm.120$ \\
Ours & .855 & .835 & .573 & $-.262\pm.013$ & .716 & .819 & .619 & $-.199\pm.022$ \\
\bottomrule
\end{tabular}
\end{table*}

\paragraph{Shortcut reversal and grading performance.}
Table~\ref{tab:supp_shortcut_qwk} shows a clear distinction between outside-FOV and inside-FOV shortcuts. For outside-FOV markers, LWM exhibits negligible QWK changes under correlation reversal on both APTOS and DDR ($|\Delta\kappa|\leq0.001$), whereas ERM and DECO experience reductions ranging from $0.157$ to $0.220$. This difference is closely associated with the LCC-based pooling mask used by LWM, rather than establishing an intrinsic advantage of prototype learning independent of preprocessing.

For inside-FOV markers, all methods remain vulnerable to reversed shortcut correlations. LWM exhibits QWK reductions of $0.262$ on APTOS and $0.199$ on DDR. Positional regularization reduces the drop relative to its ablation by $0.012$ on APTOS and $0.130$ on DDR. However, the effect is substantially more pronounced on DDR, and the ablation exhibits considerable variability across seeds ($0.120$ SD). These results indicate that positional regularization can mitigate some shortcut reliance but does not consistently prevent the exploitation of non-pathological visual cues.

\paragraph{Counterfactual prediction sensitivity.}
We quantify prediction sensitivity using $S_{\geq2}$, defined as the mean absolute change in the predicted probability of moderate-or-worse DR, $P(y\geq2)$, after inserting a marker. This threshold matches the binary severity grouping used to establish the synthetic marker-label correlation. More precisely, $S_{\geq2}=\frac{1}{N}\sum_i\left|P(y\geq2\mid I_i\oplus m)-P(y\geq2\mid I_i)\right|$, where $P(y\geq2)=\sum_{c=2}^4p_c$ and $p_c$ is the predicted class probability. We additionally report the total variation (TV) between the original and marked five-class predictive distributions, the signed change in expected severity grade ($\Delta E[y]$), and the percentage of images whose predicted grade changes.

\begin{table}[!htbp]
\centering
\caption{Counterfactual prediction sensitivity ($S_{\geq2}$) and percentage of images with changed predicted grades (Change). All configurations use three training seeds.}
\label{tab:supp_shortcut_sensitivity}
\footnotesize
\setlength{\tabcolsep}{4pt}
\renewcommand{\arraystretch}{1.08}
\begin{tabular}{lcccc}
\toprule
& \multicolumn{2}{c}{APTOS} & \multicolumn{2}{c}{DDR} \\
\cmidrule(lr){2-3}
\cmidrule(lr){4-5}
Method & $S_{\geq2}$ & Change (\%) & $S_{\geq2}$ & Change (\%) \\
\midrule
\multicolumn{5}{l}{\emph{Outside-FOV marker}} \\
ERM & .226 & 25.2 & .233 & 24.2 \\
DECO & .226 & 24.1 & .286 & 29.3 \\
Ours & .001 & 0.3 & .001 & 0.3 \\
\addlinespace[3pt]
\multicolumn{5}{l}{\emph{Inside-FOV marker}} \\
ERM & .211 & 23.5 & .246 & 25.4 \\
DECO & .248 & 26.7 & .293 & 31.2 \\
w/o Pos. Reg. & .182 & 53.2 & .172 & 70.5 \\
Ours & .180 & 52.2 & .115 & 48.2 \\
\bottomrule
\end{tabular}
\end{table}

\begin{table}[!htbp]
\centering
\caption{Counterfactual sensitivity to inside-FOV markers. $S_{\geq2}$ measures changes in the probability of moderate-or-worse DR, TV measures total variation between the five-class predictive distributions, $\Delta E[y]$ is the signed change in expected grade, and Change denotes the percentage of images with different argmax predictions. Results are averaged over three training seeds.}
\label{tab:supp_shortcut_distribution}
\footnotesize
\setlength{\tabcolsep}{4pt}
\renewcommand{\arraystretch}{1.08}
\begin{tabular}{llcccc}
\toprule
Domain & Method & $S_{\geq2}$ & TV & $\Delta E[y]$ & Change (\%) \\
\midrule
APTOS & ERM & .211 & .223 & +.313 & 23.5 \\
      & DECO & .248 & .261 & +.420 & 26.7 \\
      & Ours & .180 & .294 & +.462 & 52.2 \\
\addlinespace[2pt]
DDR & ERM & .246 & .254 & +.399 & 25.4 \\
    & DECO & .293 & .309 & +.513 & 31.2 \\
    & Ours & .115 & .219 & +.327 & 48.2 \\
\bottomrule
\end{tabular}
\end{table}

As shown in Table~\ref{tab:supp_shortcut_distribution}, $S_{\geq2}$ alone does not fully characterize prediction sensitivity. On APTOS, LWM has a smaller $S_{\geq2}$ than ERM ($0.180$ versus $0.211$), but a larger TV ($0.294$ versus $0.223$) and expected grade increase ($0.462$ versus $0.313$). This discrepancy indicates that marker-induced changes in LWM extend beyond the probability mass crossing the grade 1/2 boundary. By contrast, on DDR, LWM exhibits smaller changes than ERM and DECO across all three distribution-level measures, although $48.2\%$ of its predictions still change their argmax grade.

These observations highlight the importance of evaluating the full predictive distribution rather than relying exclusively on a single clinically relevant severity threshold. In particular, a relatively small change in $P(y\geq2)$ does not necessarily imply stable five-class grading decisions.

For outside-FOV markers, LWM exhibits negligible sensitivity under standard LCC-based pooling, with $S_{\geq2}=0.001$ on both datasets, TV of $0.002$ and $0.001$, and expected grade changes of approximately $0.001$. These results further support the effectiveness of excluding disconnected nonretinal regions from prototype pooling.

\paragraph{Diagnosing outside-FOV mask contamination.}
To explain the difference between the two marker placements, we investigate the FOV mask used for prototype pooling. The initial intensity-based mask identifies valid retinal locations using a red-channel threshold of $25/255$. Without a marker, this mask includes only $0.3\%$ of marker-region cells; inserting a white corner marker increases the fraction to $69.6\%$. Thus, the raw threshold mask can incorrectly admit disconnected nonretinal regions into prototype pooling.

We isolate this effect by retaining identical trained checkpoints and marked input images while changing only the pooling mask. We compare the threshold mask computed from the marked image, the threshold mask computed from its unmarked counterpart, and the LCC mask computed directly from the marked image. The unmarked counterpart serves only as a diagnostic control and is not required for standard inference.

\begin{table}[!htbp]
\centering
\caption{Pooling-mask intervention for outside-FOV markers. Identical checkpoints and marked inputs are used throughout. Paired: unmarked-image mask (diagnostic only); LCC: standard inference. Three seeds; $\Delta\kappa=\mathrm{Rev}-\mathrm{Corr}$.}
\label{tab:supp_shortcut_mask}
\scriptsize
\setlength{\tabcolsep}{3pt}
\renewcommand{\arraystretch}{1.06}
\begin{tabular}{llccc}
\toprule
Method & Mask & $\Delta\kappa$ & $S_{\geq2}$ & Change (\%) \\
\midrule
\multicolumn{5}{l}{\emph{APTOS: Outside-FOV}} \\
Ours & Raw & $-.245\pm.044$ & .191 & 58.9 \\
 & Paired & $-.000\pm.001$ & .001 & 0.3 \\
 & LCC & $-.000\pm.001$ & .001 & 0.3 \\
\addlinespace[2pt]
w/o Pos. & Raw & $-.269\pm.062$ & .226 & 60.4 \\
 & Paired & $-.001\pm.000$ & .001 & 0.4 \\
 & LCC & $-.000\pm.000$ & .001 & 0.5 \\
\midrule
\multicolumn{5}{l}{\emph{DDR: Outside-FOV}} \\
Ours & Raw & $-.333\pm.008$ & .197 & 76.8 \\
 & Paired & $-.001\pm.000$ & .001 & 0.3 \\
 & LCC & $-.001\pm.000$ & .001 & 0.3 \\
\addlinespace[2pt]
w/o Pos. & Raw & $-.324\pm.010$ & .191 & 78.4 \\
 & Paired & $-.001\pm.000$ & .001 & 0.3 \\
 & LCC & $-.001\pm.000$ & .001 & 0.3 \\
\bottomrule
\end{tabular}
\end{table}

As shown in Table~\ref{tab:supp_shortcut_mask}, replacing the raw threshold mask with either the paired clean mask or LCC almost entirely eliminates shortcut sensitivity. Under LCC masking, all twelve evaluated model instances exhibit negligible QWK changes, sensitivity values near $0.001$, and predicted-grade changes in at most $0.5\%$ of images. Lesion prototype responses outside the marker region change by at most $0.004\%$ at the fine scale and $2.36\%$ at the coarse scale, relative to the total lesion response over those locations. These residual changes do not meaningfully affect predictions after marker cells are excluded from pooling. On unmarked images, replacing the raw threshold mask with LCC changes clean QWK by no more than $0.006$.

This controlled intervention identifies mask contamination as the primary cause of the outside-FOV shortcut sensitivity observed with raw threshold masking. LCC filtering removes disconnected marker regions from the pooling mask without retraining, explaining the near-zero sensitivity of the standard LWM inference pipeline.

\paragraph{Discussion and limitations.}
The experiments distinguish between shortcut entry through erroneous FOV masking and shortcut reliance within the retinal region. The former is effectively mitigated by LCC-based pooling, whereas the latter remains a limitation of severity-supervised lesion prototypes. Although positional regularization improves resistance to the fixed inside-FOV marker on DDR, it does not consistently eliminate shortcut dependence across domains. More generally, constraining grading decisions to prototype-derived evidence makes the decision pathway explicit, but cannot guarantee that the learned evidence exclusively represents genuine retinal pathology.

\end{document}